%% file: main.tex
\documentclass[10pt,twocolumn,letterpaper]{article}

\usepackage[pagenumbers]{cvpr} 

\input{preamble}

\definecolor{cvprblue}{rgb}{0.21,0.49,0.74}

\usepackage[pagebackref,breaklinks,colorlinks,citecolor=cvprblue]{hyperref}
\usepackage[accsupp]{axessibility}  
\usepackage{algorithm,algorithmic}
\usepackage{bbding}
\usepackage{graphicx}
\def\paperID{6941} 
\def\confName{CVPR}
\def\confYear{2025}

\title{Multi-modal Vision Pre-training for Medical Image Analysis }

\author{
\normalsize Shaohao Rui$^{1,2*}$, Lingzhi Chen$^{2*}$, Zhenyu Tang$^{1,2}$, Lilong Wang$^{2}$, Mianxin Liu$^{2}$, Shaoting Zhang$^{2}$, Xiaosong Wang$^{2}$\Envelope\\
$^{1}$\normalsize Shanghai Jiao Tong University, Shanghai, China   $^{2}$\normalsize Shanghai AI Laboratory, Shanghai, China\\
{\tt\small\{ruishaohao, tang$_{\text{\_}}$zhenyu\}@sjtu.edu.cn}\\
{\tt\small\{chenlingzhi, wanglilong, liumianxin, zhangshaoting, wangxiaosong\}@pjlab.org.cn}\\
\href{https://github.com/openmedlab/BrainMVP}
{\texttt{https://github.com/openmedlab/BrainMVP}}
}

\begin{document}
\maketitle
\input{sec/0_abstract}    
\input{sec/1_intro}
\input{sec/2_related}
\input{sec/3_method}
\input{sec/4_Experiment}
\input{sec/5_conclusion}

\input{sec/6}
\input{sec/6_suppl}

{
    \small
    \bibliographystyle{ieeenat_fullname}
    \bibliography{main}
}

\end{document}

%% file: preamble.tex
\usepackage{indentfirst}

%% file: sec/0_abstract.tex
\begin{abstract}
Self-supervised learning has greatly facilitated medical image analysis by suppressing the training data requirement for real-world applications. Current paradigms predominantly rely on self-supervision within uni-modal image data, thereby neglecting the inter-modal correlations essential for effective learning of cross-modal image representations. This limitation is particularly significant for naturally grouped multi-modal data, e.g., multi-parametric MRI scans for a patient undergoing various functional imaging protocols in the same study. To bridge this gap, we conduct a novel multi-modal image pre-training with three proxy tasks to facilitate the learning of cross-modality representations and correlations using multi-modal brain MRI scans (over 2.4 million images in 16,022 scans of 3,755 patients), i.e., cross-modal image reconstruction, modality-aware contrastive learning, and modality template distillation. To demonstrate the generalizability of our pre-trained model, we conduct extensive experiments on various benchmarks with ten downstream tasks. The superior performance of our method is reported in comparison to state-of-the-art pre-training methods, with Dice Score improvement of 0.28\%-14.47\% across six segmentation benchmarks and a consistent accuracy boost of 0.65\%-18.07\% in four individual image classification tasks.
\end{abstract}
\renewcommand{\thefootnote}{} 
\footnotetext{
\begin{minipage}[t]{\textwidth} 
$*$~Equal contribution \\
\Envelope~Corresponding author (wangxiaosong@pjlab.org.cn)
\end{minipage}}

%% file: sec/1_intro.tex
\section{Introduction}
\label{sec:intro}
\begin{figure}[h]
    \centering
    \includegraphics[width=1.0\linewidth]{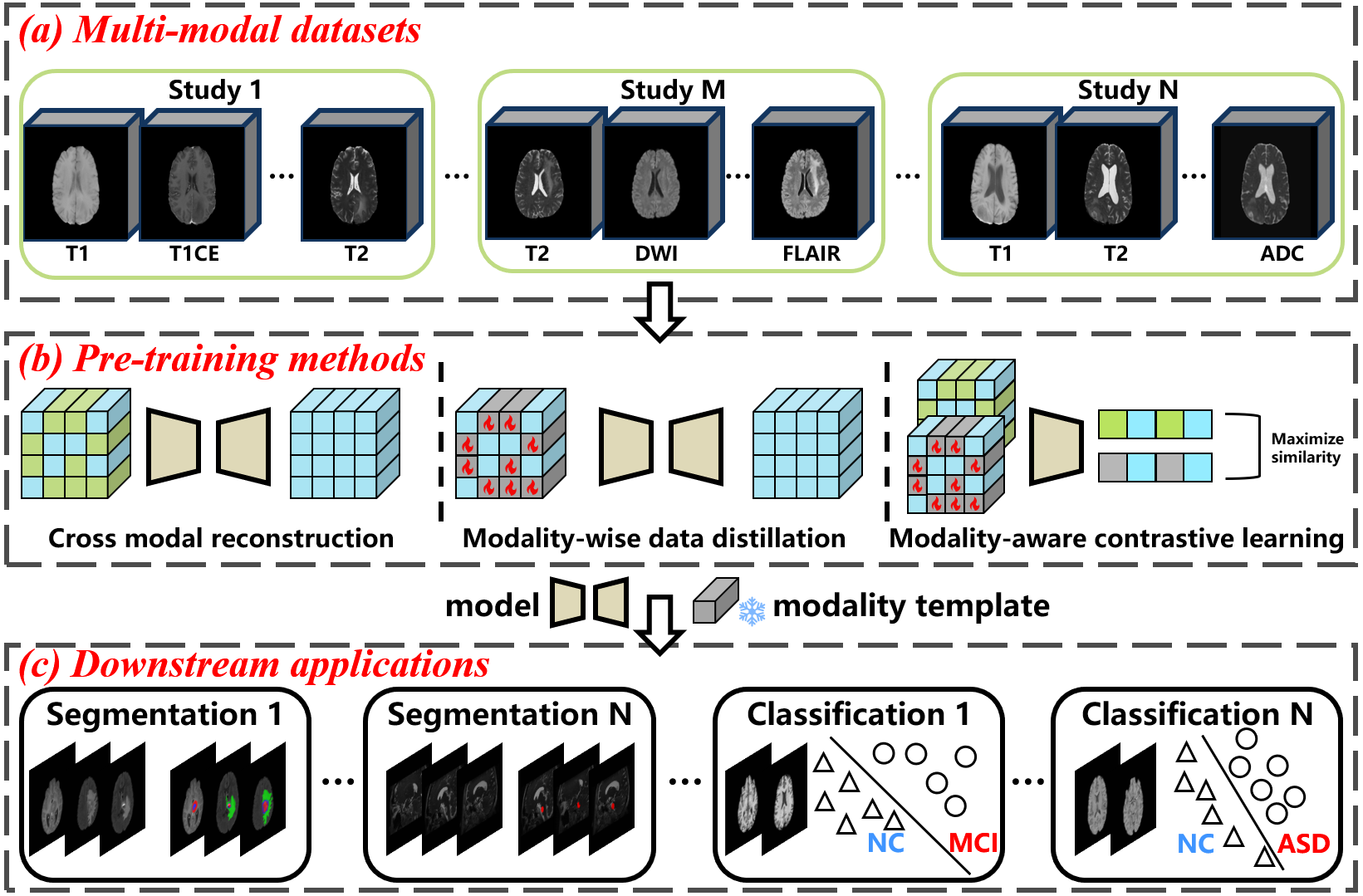}
    \caption{(a) There are naturally grouped multi-modal data, e.g., multi-parametric MRI scans in the real world. (b) We propose three proxy tasks to facilitate the learning of cross-modality representations and correlations. Blue cubes represent one modality in a study, green cubes represent another modality within the same study, and gray cubes with flame symbols represent learnable modality templates. (c) We apply the pre-trained model and distilled modality templates for downstream tasks.}
    \label{fig:preview}
\end{figure}

Medical image analysis is greatly enhanced via self-supervised learning for its capability of extracting distinctive image representation and surprisingly robust generalization performance across various downstream applications. However, current self-supervised learning methods in medical imaging are still confined to pre-training on uni-modal image data, e.g., computed tomography (CT) imaging \cite{yan2023localized,haghighi2021transferable,he2023geometric}, magnetic resonance imaging (MRI)~\cite{tang2022self,valanarasu2023disruptive,liu2023m3ae,zhou2021models}, and X-rays~\cite{cai2023dual, haghighi2022dira, zhou2021preservational,tiu2022expert,liao2022muscle}, or mixed image data (with each modality processed separately)~\cite{wu2024voco,zhang2023dive,jiang2023anatomical}. These methods mainly focus on instance-level discrimination~\cite{yan2023localized,haghighi2021transferable,wu2024voco,punn2022bt,konwer2023magnet} or image reconstruction~\cite{tang2022self,valanarasu2023disruptive,liu2023m3ae,zhou2021models} proxy tasks, failing to effectively model the relationships across modalities. In clinical practice, as shown in Fig.~\ref{fig:preview}(a), groups of multi-modal data are naturally acquired as different imaging protocols are set to capture complementary pathological features for the same patient in one examing study. 
In other words, these multi-modal data acquired on the same patient exhibit strong correspondences. For example, multi-parametric MRI (mpMRI) data combining diverse modalities helps to comprehensively depict the structural and pathological features of the brain~\cite{taleb2017self}, which substantially enhances diagnostic accuracy and thoroughness~\cite{talai2021utility}. Therefore, it is crucial to develop novel self-supervised learning frameworks for such grouped multi-modal image data, tailored to the downstream applications in the aforementioned scenarios.

In the real world, another issue of model training and inference with multi-modal data is the presence of missing modalities. Obtaining a comprehensive set of modalities in mpMRI scans can be challenging due to the complexity associated with acquisition protocol and limitations in equipment capabilities. This often leads to mismatched modality data across datasets, especially when the scale of the data amount increases dramatically. Current approaches to deal with missing modalities primarily focus on particular downstream tasks, e.g., brain tumor segmentation in BraTS~\cite{baid2021rsna,bakas2018identifying}, and have not undergone extensive investigation in large-scale cross-modal pre-training and its downstream applications~\cite{zhou2023literature, dingRFNetRegionawareFusion2021, wangMultiModalLearningMissing2023, konwerEnhancingModalityagnosticRepresentations2023, shiFTransModalityMaskedFusion2023}.

In this paper, we introduce BrainMVP, a novel self-supervised learning framework designed for multi-modal MRI data, as illustrated in Fig.~\ref{fig:preview}(b). The main goal of BrainMVP is to create generalizable cross-modal representations while effectively tackling the challenge of missing modalities during pre-training.  Thus, we also compile a dataset of 16,022 publicly available brain mpMRI scans, sourced from a range of multi-center and multi-device contributions, to demonstrate our pre-training initiatives. 

For the issue of insufficient scalability resulting from mismatched or missing modalities, we propose using uni-modal MRI inputs instead of fixed modality numbers \cite{liu2023m3ae} during pre-training. This allows for the inclusion of arbitrary numbers of modalities in the pre-training, significantly expanding the magnitude of available pre-training data. Moreover, we propose cross-modal reconstruction via masked image modeling. A key aspect of this design is the observation that different MRI modalities for the same patient often exhibit significant similarity in anatomy. By employing cross-modal reconstruction, we encourage the model to learn the disentanglement among modalities.

Towards a more generalizable pre-training model for downstream tasks, we also extract condensed structural representations of different modalities using modality-wise data distillation. Our approach is inspired by the technique of dataset distillation, which involves learning a small synthetic dataset. The performance achieved by the model training on this synthetic dataset can rival that achieved on the original large-scale datasets~\cite{wang2018dataset,zhou2022dataset,zhao2020dataset}. The learned synthetic dataset indeed encapsulates dense representations of the original dataset. In a similar idea, we optimize a set of learnable modality templates tailored for each individual modality. Intuitively, the distilled modality templates retain shared structural and statistical information about a specific modality while avoiding privacy leakage concerns associated with individual patients. Therefore, the distilled modality templates can serve as a linkage of data between pre-training and downstream tasks, i.e., as a form of information to carry and adapt between the data domains in downstream applications.

In summary, our contributions are three-fold:\par
\begin{itemize}
    \item{To the best of our knowledge, BrainMVP is the first multi-modal vision pre-training paradigm that aligns the features across modalities, targeting distinctive modality-aware representations. We also collect a dataset of 16,022 mpMRI scans (3,755 patients, over 2.4 million images) to facilitate the pre-training, covering a wide range of brain MRI scans in both diseased and healthy populations.}\par
    \item{We design two novel proxy tasks for cross-modal representation learning, i.e., cross-modal reconstruction and cross-modal contrastive learning.  To improve the generalization for downstream tasks, we also introduce the third modality-wise data distillation task to extract compact templates for each modality, benefiting both the pre-training and downstream tasks. }
    \item {We demonstrate the superior performance and the enhanced generalizability of our BrainMVP pre-trained models on ten public segmentation and classification benchmarks compared to state-of-the-art methods.}\par
\end{itemize}

%% file: sec/2_related.tex
\section{Related Work}
\label{sec:related_work}

Given the typically limited datasets available for specific medical tasks, pre-training on large-scale unlabeled data to extract highly generalizable representations is emerging as a new paradigm. Existing SSL methods in medical imaging can be roughly divided into two categories: uni-modal SSL and multi-modal SSL (with mixed modality data). While there have been numerous advancements in multi-modal learning involving paired text knowledge injection~\cite{wang2022multi,chen2022multi}, we concentrate on representation learning within medical imaging in this paper.\par
\noindent{\textbf{SSL using uni-modal data}}: 
Due to the convenience of data collection and storage, many self-supervised learning methods based on uni-modal imaging have emerged. Typical uni-modal SSL researches include computed tomography (CT) imaging\cite{yan2023localized,haghighi2021transferable,he2023geometric}, magnetic resonance imaging (MRI) ~\cite{tang2022self,valanarasu2023disruptive,liu2023m3ae,zhou2021models}, and X-rays~\cite{cai2023dual, haghighi2022dira, zhou2021preservational,tiu2022expert,liao2022muscle}. While impressive results have been achieved in specific uni-modal tasks, models pre-trained on uni-modal data often excel only in that specific modality and lack strong generalization capabilities. For example, models pre-trained on natural images struggle to generalize to medical imaging scenarios, and models trained on CT images find it challenging to generalize to MR images.
\par
\noindent{\textbf{SSL using mixed modality data}}:
It has been validated that multi-modal data from different imaging sources can be unified through shared encoders in a self-supervised learning manner and also play a complementary role in promoting the representation learning of specific modalities~\cite{xie2024refs, valanarasu2023disruptive, liu2023m3ae, taleb2021multimodal}. Composing CT, X-ray, and MR images, PCRLv2~\cite{zhou2023unified} addresses the issue of local information loss in medical images within the contrastive learning SSL paradigm by suggesting pixel recovery and feature alignment at various scales for diverse enhancement samples. Additionally, PCRLv2~\cite{zhou2023unified} recommends implementing SSL without using skip connections to avoid shortcut solutions in pixel restoration. Using CT and X-ray images, VoCo~\cite{wu2024voco} leverages the contextual position priors to learn consistent semantic representations in pre-training and performs exceptionally well in medical images where the relative positions are relatively fixed. 
Although the methods above involve joint training on multi-modal data, different data sources often pose a bottleneck to the model's cross-modal understanding. \cite{taleb2017self} introduces a multi-modal puzzle task designed to enhance representation learning from various image modalities
and applies modal transformation based on a generative network while solely acting as a data augmentation strategy. 
Our method, instead, employs a simple yet effective strategy of cross-modal reconstruction to learn cross-modal representations, incorporating modal complementary properties into the pre-training proxy task. 
\par
\noindent{\textbf{Dataset distillation for knowledge compression}}:
 Dataset distillation is first proposed to distill a core set in which the learned model can achieve a performance comparable to that of the whole dataset~\cite{wang2018dataset}. In this way, computational burden and data storage costs can be significantly reduced~\cite{zhao2020dataset}. Existing dataset distillation methods can mainly be categorized into three types: parameter matching~\cite{zhao2020dataset,zhao2021dataset,jiang2023delving,cazenavette2022dataset,li2024dataset,cui2023scaling}, distribution matching~\cite{zhao2023dataset,wang2022cafe} and performance matching~\cite{deng2022remember,nguyen2021dataset,loo2022efficient}. 
 The novel modality data distillation method presented in this paper is inspired by performance matching methods, where distilled templates are learned via reconstructing the real modality image along the pre-training trajectories. 
This approach first learns patient-agnostic structural representations within modalities and then integrates patient-specific modality information during downstream tasks to bridge the domain gap and enhance model generalization.

%% file: sec/3_method.tex
\section{Methods}
\begin{figure*}
    \centering
    \includegraphics[width=1\linewidth]{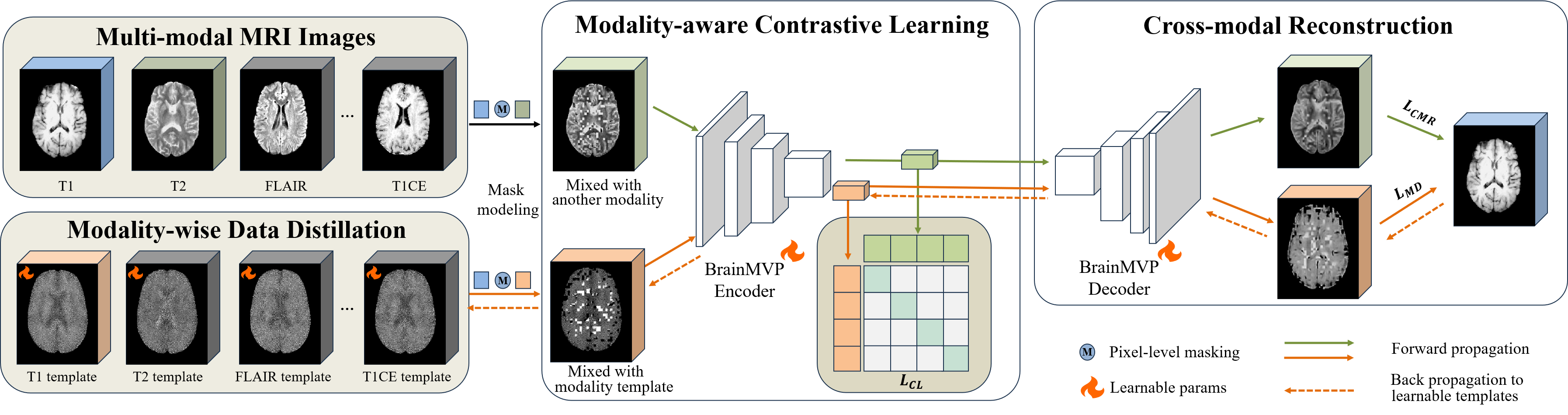}
    \caption{Overview of the proposed BrainMVP, comprised of (a) cross-modal reconstruction module that aims at learning a mapping from images masked with another modality to the original; (b) modality-wise data distillation module that learns condensed modality templates via gradient backpropagation; and (c) modality-aware contrastive learning module for introducing study/case-level modality invariance to the learned features. }
    \label{fig:overview}
\end{figure*}

As shown in Fig.~\ref{fig:overview}, BrainMVP comprises three key modules: cross-modal reconstruction, modality-wise data distillation, and modality-aware contrastive learning. 
Three modules are detailed in the following sections.
\subsection{Cross-Modal Reconstruction}
\label{sec:Cross-Modal Reconstruction}
\noindent{\textbf{Problem Setting}}: Given an unlabeled dataset 
$\mathcal{D}=\{X_{im}\in\mathbb R^{D\times H\times W}|m\in \{1,\dots,M_{i}\}, i\in\{1,\dots,N\}\}$, where $M_{i}$ denotes the number of modalities in the $i$-th sample and $N$ represents total number of samples. Masked image modeling (MIM) first masks with noise or discards (denoted as $\Phi(\cdot)$ ) a large portion of $X_{im}$ to obtain a masked input $\Phi({X}_{im})$, and then reconstructs the original image from it to learn efficient representations. 
Specifically, let the model be $\mathcal F(\cdot)=\mathcal F_{dec}\circ \mathcal F_{enc}(\cdot)$, where $\mathcal F_{\rm enc}(\cdot)$ and $\mathcal F_{\rm dec}(\cdot)$ are the encoder and decoder respectively,
MIM minimizes the following reconstruction loss:
\begin{equation}
\mathcal L_{rec}=\left|\left|\mathcal F_{\rm dec}(\mathcal F_{\rm enc}(\Phi({X}_{im}))-X_{im}\right|\right|_2.
\end{equation}
The core idea of our proposed reconstruction proxy tasks, which are elaborated in Sections~\ref{sec:Cross-Modal Reconstruction} and \ref{sec:Modality-wise Data Distillation}, is to obtain meaningful representations via exploiting different forms of $\Phi(\cdot)$ function.
\par
\noindent{\textbf{Pixel-level cross-modal masking.}} Given a uni-modal input volume $X_{im}$ sampled from an mpMRI case (with $M_{i}$ modalities), cross-modal masking aims to mask out a large region of $X_{im}$ and replace with another modality image $X_{in}$ (also sampled from $X_{i}$, $n\neq m$). Specifically, we first randomly mask a region of size $r\times r \times r$ in $X_{im}$, where $r$ denotes the size of each dimension of 3D volumes. Then, we fill in the masked region with a patch cropped with the same location and size on another modality of the sampled case. Finally, we repeat the above masking-filling operation until the proportion of masked pixels over the total input volume ($X_{im}$) pixels arrives $p^*$. More details of the masking algorithm can be found in the supplementary materials. \par

\noindent{\textbf{Cross-modal reconstruction.}} Let our proposed cross-modal masking strategy be $\Phi_{modal}$. Given that the masking operation masks a large portion of the image, the resulting masked input volume $\Phi_{modal}(X_{im}, X_{in})$ will contain information predominantly from $X_{in}$. The extracted representation $\mathcal F_{enc}(\Phi_{modal} (X_{im}, X_{in}))$ will thus encode a significant amount of semantic information from $X_{in}$. Since we do not introduce skip connections between the encoder and decoder, we only reconstruct $X_{im}$ from the latent representation $\mathcal F_{enc}(\Phi_{modal}(X_{im}, X_{in}))$, which is a challenging task for natural images. However, due to the high structural similarity between different modalities in mpMRI data, with strong contrasts only in certain regions, the cross-modal reconstruction can encourage the model to learn cross-modal representations and explore the correlations between different modalities. Formally, the cross-modal reconstruction loss can be expressed as:
\begin{align}
\resizebox{0.9\linewidth}{!}{
$
\mathcal L_{CMR}=\left|\left|\mathcal F_{dec}(\mathcal F_{enc}(\Phi_{modal}({X}_{im}, X_{in}))-X_{im}\right|\right|_2
$.
}
\end{align}

\subsection{Modality-wise Data Distillation}
\label{sec:Modality-wise Data Distillation}
The primary objective of the foundation model is to extract highly generalizable latent representations. However, the proxy tasks currently used in pre-training models are often unrelated to the downstream application tasks. We attempt to introduce certain bridging components during the pre-training stage that can guide the pre-training process in acquiring the necessary specific representations. Simultaneously, we hope that these bridging components can facilitate the feature expression of the pre-trained model when applied to downstream tasks. As shown in Fig.~\ref{fig:overview}, modality-wise data distillation is in conjunction with the cross-modal reconstruction process. Specifically, in the cross-modal reconstruction part, we use data either from another modality image $X_{in}$ to fill in the masked region in $X_{im}$ or from the corresponding learnable modality template.

Specifically, the learnable modality templates $T=\{T_{m}\}_{m=1}^S$ sized $S \times H \times W \times D$ are initialized with zero, where $S$ represents the number of modalities in the pre-training datasets. Similar to cross-modal reconstruction, the image needed for filling in $X_{im}$ is $T_{m}$ ($m$ represents the corresponding modality) instead of another modality, and the remaining steps are the same. An example of learned modality templates is shown in Fig.~\ref{fig:template_1}, which demonstrates a compact representation of the structural information for each modality along the pre-training trajectories.
Given the masking strategy for modality-wise data distillation denoted as $\Phi_{distill}$, the corresponding loss can be expressed as:
\begin{align}
\mathcal L_{MD}=\left|\left|\mathcal F_{dec}(\mathcal F_{enc}(\Phi_{distill}({X}_{im}, T_m))-X_{im}\right|\right|_2.
\end{align}\par
Cross-modal reconstruction and modality-wise data distillation are performed simultaneously. The model needs to learn not only the structural information of a specific modality to form the distilled modality templates but also the transformation relationship between modalities.  The representations learned by our pre-trained model are considered modality-agnostic and contain fused representations of different modalities.

\subsection{Modality-aware Contrastive Learning}
\label{sec:Modality-aware Contrastive Learning}
As described in section \ref{sec:Cross-Modal Reconstruction} and \ref{sec:Modality-wise Data Distillation}, $\Phi_{modal}(X_{im}, X_{in})$ and $\Phi_{distill}(X_{im}, T_m)$ still contain $(1-p^*)$ proportion of information about $X_{im}$, we aim to keep feature-level consistency. To this end, we use contrastive loss to close the high-dimension feature discrepancy.
Given their partial semantic consistency, $\Phi_{modal}(X_{im}, X_{in})$ and $\Phi_{distill}(X_{im}, T_m)$ form positive pairs.
This can be formalized as:
\begin{equation}
\mathcal{L}_{f_{im}\to g_{im}} = -\log\frac{\exp(f_{im} \cdot g_{im}^{T}/\tau  )}{\sum_{j=1}^{|\mathcal B|} \exp(f_{im} \cdot g_{jm}^{T}/\tau)}, 
\end{equation}
where $f_{im}$ represents the embedding from the current modality image masked with another in the same study, while $g_{im}$ represents the embedding from the current modality image masked with the corresponding distilled template. $|\mathcal B|$ denotes the number of positive pairs in a batch.
The loss $\mathcal{L}_{g_{im}\to f_{im}}$ is calculated by swapping $f_{im}$ and $g_{im}$. The total loss is the sum of both terms:
\begin{equation}
\mathcal{L}_{CL} = \frac{1}{2}\left(\mathcal{L}_{f_{im}\to g_{im}} + \mathcal{L}_{g_{im}\to f_{im}}\right). 
\end{equation}
\par
\vspace{1.2mm}
\textbf{Overall loss.}
In summary, the total loss for the proposed multi-modal self-supervised learning scheme is a combination of $\mathcal L_{CMR}$, $\mathcal L_{MD}$, and $\mathcal L_{CL}$:
\begin{equation}
\begin{aligned}
    \mathcal L_{SSL}=\frac{1}{|\mathcal B|}\sum_{\mathrm{i}=1}^{|\mathcal B|}\frac{1}{M_{i}}\sum_{m=1}^{M_{i}}(\mathcal L_{CMR}&+\lambda_{MD}\cdot \mathcal L_{MD}\\
    &+\lambda_{CL}\cdot \mathcal L_{CL}),
\end{aligned}
\end{equation}
where $\lambda_{MD}$ as well as $\lambda_{CL}$ are coefficients for balancing the corresponding loss term contributions, and $M_{i}$ denotes the number of modalities of study/case $i$.

\subsection{Modality templates for downstream application}
The distilled modality templates carry shared structural representations of specific modalities from pre-training datasets. We aim to apply these templates to downstream tasks for enhancing generalization performance. In essence, we randomly replace the multi-modal MRI scans of downstream tasks with corresponding distilled templates, aiming to improve the model's modality-invariant representation learning. For detailed implementation, please refer to the supplementary materials.

%% file: sec/4_Experiment.tex
\section{Experiments}
\subsection{Datasets}

\noindent{\textbf{Pre-training datasets}}:
We curate a large-scale pre-training dataset collected from five publicly available mpMRI datasets with various sites and acquisition protocols, spanning eight modalities with a total of 3,755 cases/patients containing 16,022 3D scans as demonstrated in Table~\ref{table:dataset}. 
Among them, the BraTS2021~\cite{baid2021rsna}, BraTS2023-SSA~\cite{adewole2023brain}, and BraTS2023-MEN~\cite{labella2023asnr} datasets encompass four common multi-modal brain MRI scans, including T1, T1CE, T2, and FLAIR. The UCSF-PDGM~\cite{calabrese2022university} dataset includes 501 cases with additional DWI and ADC modalities. The IXI\footnote{https://brain-development.org/ixi-dataset/} dataset is utilized as a supplement to the pre-training dataset with cases of normal brains, also incorporating richer modalities such as MRA and PD. Notably, we do not use the segmentation annotations provided in these datasets, and the data do not overlap with the test sets of downstream tasks. More details are in the supplementary materials.\par
\noindent{\textbf{Downstream datasets}}:
Ten datasets with various MR imaging sequences are employed for evaluation. These include segmentation tasks: (1) pediatric tumor segmentation BraTS2023-PED~\cite{kazerooni2023brain}; (2) brain metastases segmentation BraTS2023-MET~\cite{moawad2023brain}; (3) ischemic stroke lesion segmentation ISLES22~\cite{hernandez2022isles}; (4) brain structure segmentation MRBrainS13~\cite{mendrik2015mrbrains}; (5) gliomas segmentation UCSF-PDGM~\cite{calabrese2022university}; (6) vestibular schwannoma segmentation VSseg~\cite{shapey2021segmentation};
and classification tasks: (1) high-grade and low-grade glioma classification BraTS2018~\cite{bakas2018identifying}; (2) mild cognitive impairment classification ADNI~\cite{jack2008alzheimer}; (3) attention deficit hyperactivity disorder classification ADHD-200~\cite{adhd2012adhd}; (4) autism spectrum disorder classification ABIDE-I~\cite{di2014autism}. More details about these downstream datasets are in the supplementary materials.\par
\begin{table}[!htbp]
 \resizebox{0.48\textwidth}{!}{
\begin{tabular}{lllll}
\hline
Dataset  & Task type     & Modality type    & cases  \\ \hline
\textbf{\textit{Pre-training}}    &  &  &   3755&\\ 
BraTS2021~\cite{baid2021rsna}& -  & T1,T1CE,T2,FLAIR   & 1470   \\ 
BraTS2023-SSA~\cite{adewole2023brain}  & - & T1,T1CE,T2,FLAIR   & 75  \\ 
BraTS2023-MEN~\cite{labella2023asnr}  &-   & T1,T1CE,T2,FLAIR   & 1141     \\ 
UCSF-PDGM~\cite{calabrese2022university}&-   & T1,T1CE,T2,FLAIR,DWI,ADC & 501      \\ 
IXI\footnote{https://brain-development.org/ixi-dataset/}    &-   & T1,T2,MRA,PD & 568     & -  \\ 
\hline

\textbf{\textit{Downstream}}     &   &  &   &\\ 
BraTS2023-PED~\cite{kazerooni2023brain}  & seg. (pediatric tumor)  & T1,T1CE,T2,FLAIR   & 99   \\ 
BraTS2023-MET~\cite{moawad2023brain}  & seg. (brain metastases)  & T1,T1CE,T2,FLAIR   & 238      \\ 
ISLES22~\cite{hernandez2022isles}& seg. (ischemic stroke lesion)  & FLAIR,DWI,ADC& 238      \\ 
MRBrainS13~\cite{mendrik2015mrbrains}     & seg. (CF,GM,WM)  & T1,T1CE,FLAIR  & 20   \\ 
UPENN-GBM~\cite{bakas2021multi}& seg. (glioblastoma)  & T1,T1CE,T2,FLAIR   & 127      \\ 
VSseg~\cite{shapey2021segmentation}    & seg. (vestibular schwannoma)  & T1    & 242    \\ 
BraTS2018~\cite{bakas2018identifying} & cls. (HGG and LGG) & T1,T1CE,T2,FLAIR   & 285      \\ 
ADNI~\cite{jack2008alzheimer}     & cls. (MCI and NC) & T1     & 1348  \\ 
ADHD-200~\cite{adhd2012adhd}     & cls. (ADHD and NC) & T1     & 767    \\ 
ABIDE-I~\cite{di2014autism}    & cls. (ASD and NC) & T1     & 819    \\ \hline
\end{tabular}
}
\caption{Details of datasets used in our work. seg.: segmentation; cls.: classification; CF: Cerebrospinal Fluid; GM: Gray Matter; WM: White Matter; HGG: Higher Grade Glioma; LGG: Lower Grade Glioma; MCI: Mild Cognitive Impairment; NC: Normal Control; ADHD: Attention Deficit Hyperactivity Disorder; ASD: Autism Spectrum Disorder.}
\label{table:dataset}
\end{table}

\begin{table*}[t]
  \setlength{\tabcolsep}{2.5pt} %
 \resizebox{1.0\textwidth}{!}{
\begin{tabular}{ccccccccccccccccccccc}
\hline
\textbf{Method}               & \textbf{Modality}        & \textbf{Network}    & \multicolumn{4}{c}{\textbf{BraTS2023-PED}~\cite{kazerooni2023brain}} & \multicolumn{4}{c}{\textbf{BraTS-MET}~\cite{moawad2023brain}} & \textbf{ISLES22~\cite{hernandez2022isles}} & \multicolumn{4}{c}{\textbf{MRBrainS13}~\cite{mendrik2015mrbrains}}                         & \textbf{VSseg~\cite{shapey2021segmentation}} & \multicolumn{4}{c}{\textbf{UPENN-GBM}~\cite{bakas2021multi}} \\ \hline
                     & &   & ET  & TC     & WT     & AVG    & ET      & TC     & WT     & AVG    & IS & CF & GM & WM & AVG   & VS        & ET      & TC     & WT     & AVG    \\ \hline
\multicolumn{1}{c}{\textbf{\textit{From Scratch}}}       &            &        &        &        &        &         &        &        &        &              &                     &             &                &       &            &         &        &        &        \\
UNETR~\cite{hatamizadeh2022unetr}                & -               & -          & 46.46  & 76.43  & 78.66  & 67.19  & 54.01   & 54.87  & 59.44  & 56.11  & 74.65        & 67.55               & 78.73       & 83.69          & 76.66 & 70.28      & 83.10   & 80.88  & 81.98  & 81.99  \\
UNET3D~\cite{ronneberger2015u}               & -               & -          & 47.12  & 81.60  & 83.94  & 70.89  & 56.44   & 58.75  & 62.76  & 59.32  & 80.94        & 70.47               & 73.93       & 82.96          & 75.78 & 69.43      & 85.65   & 88.76  & 86.27  & 86.89  \\
UniFormer~\cite{li2023uniformer}            & -               & -          & 46.73  & 83.87  & 86.97  & 72.52  & 67.22   & \underline{72.74}  & \underline{70.78}  & 70.25  & 84.97        & 77.66               & 74.09       & 75.60          & 75.78 & \underline{80.33}      & \underline{87.93}   & 91.86  & 88.81  & 89.53  \\
Swin-UNETR~\cite{hatamizadeh2021swin}           & -               & -          & 49.66  & 81.10  & 84.13  & 71.63  & 63.84   & 67.08  & 68.58  & 66.50  & 75.88        & 70.35               & \underline{81.66}       & 84.65          & 78.89 & 76.82      & 87.60   & 91.15  & 87.34  & 88.70  \\
\hline
\multicolumn{1}{c}{\textbf{\textit{With General SSL}}} &            &        &        &        &        &         &        &        &        &              &                     &             &                &       &            &         &        &        &        \\
MAE3D~\cite{he2022masked,chen2023masked}               & Natural         & UNETR      & 46.55  & 77.08  & 79.32  & 67.65  & 57.45   & 59.19  & 62.06  & 59.57  & 70.43        & 68.30               & 80.57       & 84.69          & 77.86 & 69.57      & 83.66   & 80.42  & 81.86  & 81.98  \\
SimMIM~\cite{xie2022simmim}               & Natural         & UNETR      & 45.14  & 76.59  & 78.61  & 66.78  & 54.46   & 55.84  & 58.89  & 56.40  & 69.94        & 68.11               & 80.49       & \underline{84.76}          & 77.79 & 69.08      & 83.70   & 81.68  & 82.44  & 82.61  \\
MoCov3~\cite{chen2021empirical}             & Natural         & UNETR      & 45.66  & 77.37  & 79.88  & 67.64  & 55.84   & 56.77  & 61.62  & 58.07  & 70.32        & 67.97               & 79.64       & 84.36          & 77.32 & 69.83      & 83.02   & 80.54  & 81.77  & 81.78  \\
\hline
\multicolumn{1}{c}{\textbf{\textit{With Medical   SSL}}} &            &        &        &        &        &         &        &        &        &              &                     &             &                &       &            &         &        &        &        \\
MG~\cite{zhou2021models}                   & CXR, CT        & UNET3D     & 47.99  & \textbf{86.69}  & \underline{88.41}  & \underline{74.36}  & 60.11   & 64.05  & 65.43  & 63.19  & 83.53        & 71.40               & 74.71       & 80.41          & 75.51 & 76.33      & 86.64   & 90.58  & 87.03  & 88.08  \\
TransVW~\cite{haghighi2021transferable}              & CT              & UNET3D     & 46.38  & 80.05  & 81.98  & 69.47  & 56.10   & 58.69  & 62.81  & 59.20  & 80.24        & 68.92               & 80.53       & 83.70          & 77.72 & 71.76      & 85.95   & 89.51  & 86.91  & 87.46  \\
GVSL~\cite{he2023geometric}                 & CT              & UNET3D     & 49.05  & 84.47  & 86.81  & 73.45  & 62.46   & 66.81  & 67.26  & 65.51  & 80.05        & 69.34               & 75.07       & 82.85          & 75.75 & 72.21      & 87.09   & 91.75  & 87.53  & 88.79  \\
Swin-UNETR*~\cite{tang2022self}          & MRI             & Swin-UNETR & 49.07  & 81.74  & 84.13  & 71.65  & 60.60   & 64.56  & 64.53  & 63.23  & 79.55        & 69.67               & \textbf{82.09}       & \textbf{86.13}          & \underline{79.30} & 75.55      & 87.24   & 91.46  & 87.28  & 88.66  \\
VoCo~\cite{wu2024voco}                 & MRI             & Swin-UNETR & 48.66  & 82.26  & 84.64  & 71.85  & 57.49   & 59.33  & 63.59  & 60.13  & 77.58        & 71.29               & 76.43       & 81.40          & 76.37 & 76.45      & 86.65   & 90.54  & 87.34  & 88.18  \\
DAE~\cite{valanarasu2023disruptive}                  & MRI             & Swin-UNETR & 49.30  & 82.12  & 84.78  & 72.07  & 62.27   & 65.99  & 64.85  & 64.37  & 73.92        & 71.37               & 78.50       & 83.20          & 77.69 & 74.51      & 86.90   & 90.83  & 87.32  & 88.35  \\
M$^3$AE~\cite{liu2023m3ae}                 & MRI             & UNET3D     & 46.77  & 85.67  & 86.89  & 73.11  & 66.01   & 70.92  & 70.18  & 69.04  & 83.85        & 71.32               & 69.56       & 79.28          & 73.39 & 75.96      & 87.15   & 91.90  & 88.44  & 89.16  \\
M$^3$AE~\cite{liu2023m3ae}                  & MRI             & UniFormer  & \underline{50.77}  & 84.95  & 86.70  & 74.14  & \underline{68.08}   & 72.35  & 70.74  & \underline{70.39}  & \underline{86.32}        & \underline{78.23}               & 77.20       & 76.43          & 77.29 & 79.31      & 87.75   & \underline{92.43}  & 88.72  & \underline{89.63}  \\
\textbf{BrainMVP}         & MRI             & UNET3D     & 47.75  & 85.99  & \textbf{88.46}  & 74.07  & 67.24   & 71.27  & 68.63  & 69.05  & 83.31        & 68.88               & 74.60       & 82.66          & 75.38 & 76.02      & 87.30   & 91.87  & \underline{88.98}  & 89.38  \\
\textbf{BrainMVP}         & MRI             & UniFormer  & \textbf{55.45} & \underline{86.54}  & \underline{88.41}  & \textbf{76.80}  & \textbf{70.70}   & \textbf{75.80}  & \textbf{74.52}  & \textbf{73.67}  & \textbf{86.60}        & \textbf{81.04}               & 78.17       & 81.61          & \textbf{80.27} & \textbf{83.64}      & \textbf{88.49}   & \textbf{92.48}  & \textbf{89.07}  & \textbf{90.01} \\
\hline
\end{tabular}}
\caption{Experimental results on six downstream \textbf{segmentation} datasets. We report the mean Dice score (\%) on each dataset and the best results are bolded. The second best results are underlined. CXR: Chest X-Ray; ET: enhancing tumor; TC: tumor core; WT: whole tumor; AVG:average; IS: Ischemic Stroke; CF: Cerebrospinal Fluid; GM: Gray matter; WM: White matter; VS: Vestibular schwannoma.}
\label{table:seg_res}
\end{table*}
\subsection{Implementation details.} 
\begin{figure}
    \centering
    \includegraphics[width=1.0\linewidth]{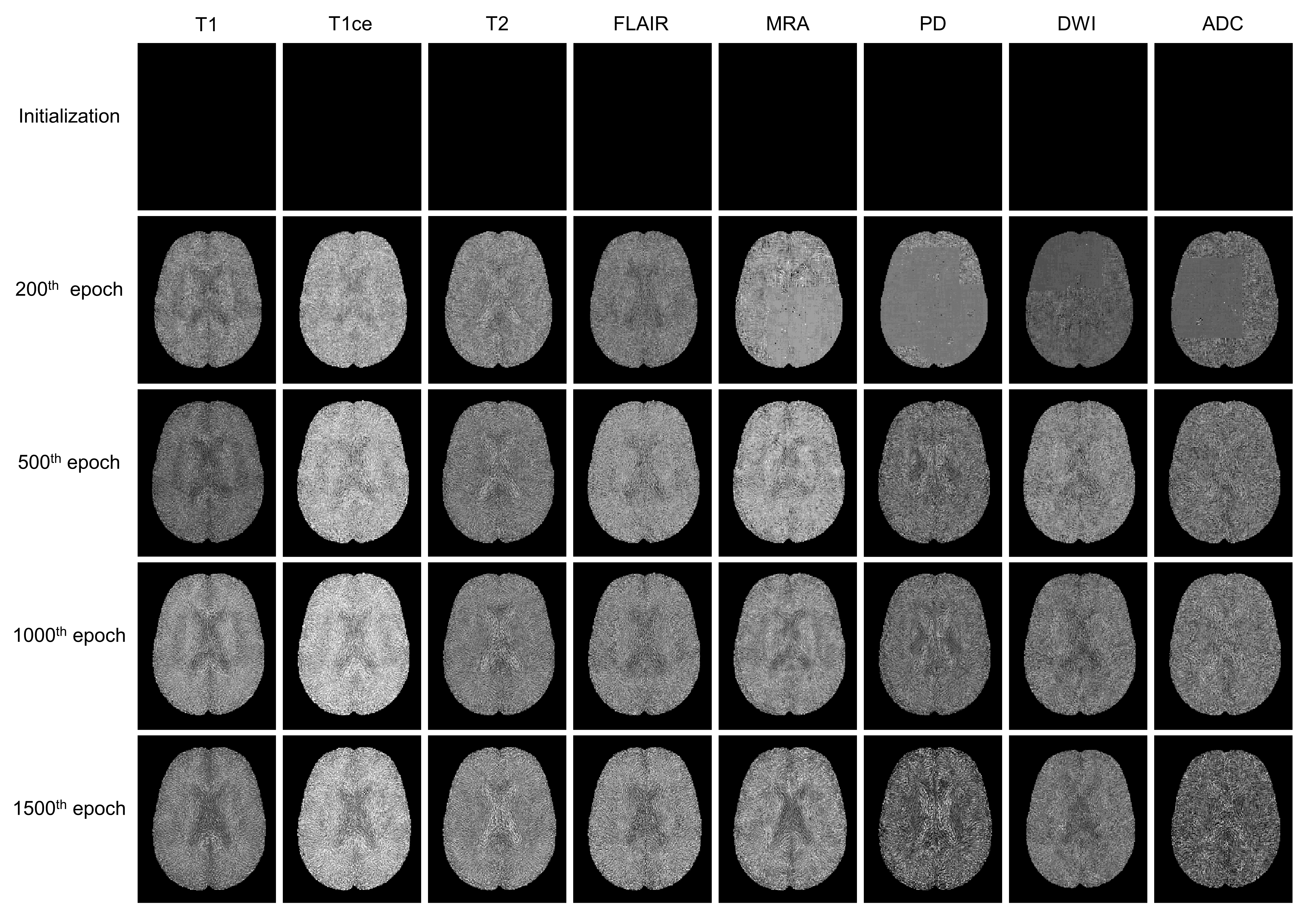}
    \caption{Visualization of distilled modality templates along the pre-training trajectories.}
    \label{fig:template_1}
\end{figure}

We adopt UniFormer~\cite{li2023uniformer} as the backbone of BrainMVP due to its natural multi-modal fusion capabilities.
In addition, we have conducted experiments based on the UNET3D~\cite{ronneberger2015u} network as well. 
All the experiments are implemented with PyTorch and are run on 8 NVIDIA GeForce RTX 4090 GPUs. Referring to~\cite{liu2023m3ae}, we set $r=8$ and $p^*=0.875$. $\lambda_{MD}$ and $\lambda_{CL}$ are both set to 1.0 for equal treatment. During pre-training, we use the AdamW~\cite{loshchilov2017decoupled} optimizer with a momentum of 0.9 and the weight decay is 1e-5. We train the model for 1,500 epochs with a batch size of 3 and introduce the modality-aware contrastive learning module at epoch 1000 (when the distilled templates have been trained visually well and the corresponding loss has converged, shown in Fig.\ref{fig:template_1}). The initial learning rate is set to 3e-4 and we employ a cosine learning rate decay strategy. Detailed hyperparameters for downstream experiments can be found in the supplementary materials.
\par
\noindent{\textbf{Comparison methods}}. We compare our BrainMVP against three different types of approaches, i.e., training from scratch, general domain SSL methods, and medical domain SSL methods. There are three mainstream medical image segmentation networks for training from scratch: UNETR~\cite{hatamizadeh2022unetr}, UNET3D~\cite{ronneberger2015u}, and Swin-UNETR~\cite{hatamizadeh2021swin}. UniFormer~\cite{li2023uniformer} is a novel 3D medical image segmentation network initially developed in the field of video object detection and extensive experiments have been conducted to verify its effectiveness. The subsequent SSL methods are pre-trained on the above architectures, allowing for a fair comparison of the impact of different network architectures on the final performance. The baseline SSL methods include MAE3D~\cite{he2022masked,chen2023masked}, MIM-based SimMIM~\cite{xie2022simmim}, and contrastive learning related MoCoV3~\cite{chen2021empirical} for general domain, and MG~\cite{zhou2021models}, TransVW~\cite{haghighi2021transferable}, GVSL~\cite{he2023geometric}, Swin-UNETR~\cite{tang2022self}, and VoCo~\cite{wu2024voco} for medical domain. Specifically, two MIM-based methods in medical domain, namely, DAE~\cite{valanarasu2023disruptive} and M$^3$AE~\cite{liu2023m3ae}, are also taken for comparison. For MRI modality, we \textbf{re-implement} the baseline methods on our pre-training dataset for a fair comparison.  \par
\noindent{\textbf{Label efficiency experiments}}. To validate if our BrainMVP, pre-trained on large-scale mpMRI datasets, can significantly reduce annotation workload in clinical practice, particularly for handling label-deficient segmentation tasks (which incur higher annotation costs), we conduct label efficiency experiments on five segmentation and one classification datasets. Specifically, we randomly split the training labeled samples into five partitions and gradually increase the training set size by one partition at a time until reaching the full dataset size. The resulted experiments are configured with 20\%, 40\%, 60\%, 80\%, and 100\% of the total training data. The validation and test sets are kept the same for a fair comparison. For the comparison methods, we select representative approaches for each pre-training data modality (natural, CT, and MRI), including MAE3D ~\cite{he2022masked,chen2023masked}, GVSL~\cite{he2023geometric}, MG~\cite{zhou2021models}, and VoCo~\cite{wu2024voco}. Notably, we observe that MG~\cite{zhou2021models} exhibits strong generalization performance across many datasets, so we include it for comprehensiveness of the comparison. \par
\noindent{\textbf{Evaluation metrics}}. For segmentation tasks, we use Dice Score and Hausdorff distance at 95th percentile (HD95) as evaluation metrics. For classification tasks, we report accuracy (ACC), area under the curve (AUC), and F1 score for comprehensive assessment with higher metric values indicating better classification performance. Note that the HD95 results and qualitative experimental results are presented in the supplementary materials.

\subsection{Experiments on downstream tasks}\par
\noindent{\textbf{Superior performance on tumor segmentation datasets.}} We first validate our BrainMVP on BraTS2023-PED~\cite{kazerooni2023brain} and UPENN-GBM~\cite{bakas2021multi}. As shown in Table~\ref{table:seg_res}, medical-specific SSL methods consistently outperform general SSL approaches, as models pre-trained on natural images generalize poorly to medical imaging. Specifically, the best average Dice Score achieved by general SSL methods based on MIM is 67.65\%, which is 9.15\% lower than BrainMVP's best result of 76.80\%. Also, MoCoV3~\cite{chen2021empirical} performs less effectively, achieving 9.16\% lower in Dice Score compared to BrainMVP. This disparity arises because typical pre-training methods developed primarily for 2D image tasks often require full images or large patches as input, which is usually impractical for 3D medical images. Our BrainMVP also outperforms medical SSL methods based on mask modeling, such as M$^3$AE~\cite{liu2023m3ae} (76.80\% \textit{\textit{vs.}} 74.14\%) and DAE~\cite{valanarasu2023disruptive} (76.80\% \textit{\textit{vs.}} 72.07\%). We further validate the effectiveness of BrainMVP on UPENN-GBM~\cite{bakas2021multi}, as shown in Table~\ref{table:seg_res}. BrainMVP achieves an average Dice Score of 90.01\% and outperforms state-of-the-art methods.
\begin{table*}[t]
 \resizebox{1.0\textwidth}{!}{
\begin{tabular}{ccccccccccccccccccccc}
\hline
\textbf{Method}               & \textbf{Modality}        & \textbf{Network}    & \multicolumn{3}{c}{\textbf{BraTS2018~\cite{bakas2018identifying}}} & \multicolumn{3}{c}{\textbf{ADNI}~\cite{jack2008alzheimer}} & \multicolumn{3}{c}{\textbf{ADHD-200~\cite{adhd2012adhd}}} & \multicolumn{3}{c}{\textbf{ABIDE-I~\cite{di2014autism}}}
 \\
\hline
                     &                 &            & ACC         & AUC        & F1         & ACC    & AUC    & F1     & ACC    & AUC    & F1     & ACC     & AUC    & F1      \\
\hline
\multicolumn{1}{c}{\textbf{\textit{From Scratch}}}       &            &             &            &            &        &        &        &        &        &        &         &        &         \\
UNETR~\cite{hatamizadeh2022unetr}                & -               & -          & 0.7895      & 0.7817     & 0.6621     & 0.5672 & 0.6066 & 0.5645 & 0.6688 & 0.6523 & 0.6204 & 0.6121  & 0.5478 & 0.5507  \\
UNET3D~\cite{ronneberger2015u}               & -               & -          & 0.7368      & 0.7373     & 0.4242     & 0.5756 & 0.4966 & 0.3653 & 0.6494 & 0.6798 & 0.4265 & 0.6061  & 0.5059 & 0.4591  \\
UniFormer~\cite{li2023uniformer}            & -               & -          & 0.7762      & 0.7719     & 0.6994     & 0.5546 & 0.6343 & 0.5526 & 0.6039 & 0.6387 & 0.5796 & 0.5879  & 0.4433 & 0.4292  \\
Swin-UNETR~\cite{hatamizadeh2021swin}           & -               & -          & 0.7018      & 0.7143     & 0.6069     & 0.5672 & 0.5853 & 0.5650 & 0.6494 & 0.6950 & 0.6240 & 0.6121  & 0.5530 & 0.5596  \\
\hline
\multicolumn{1}{c}{\textbf{\textit{With General   SSL}}} &            &             &            &            &        &        &        &        &        &        &         &        &         \\
MAE3D~\cite{he2022masked,chen2023masked}               & Natural         & UNETR      & 0.7018      & 0.6754     & 0.5645     & 0.5756 & 0.5414 & 0.5651 & 0.6169 & 0.6489 & 0.5906 & 0.6061  & 0.4983 & 0.4591  \\
SimMM~\cite{xie2022simmim}                   & Natural         & UNETR      & 0.7368      & 0.8349     & 0.7077     & 0.6218 & 0.6026 & 0.5446 & 0.6234 & 0.6567 & 0.5790 & 0.5394  & 0.5819 & 0.5318  \\
MoCov3~\cite{chen2021empirical}               & Natural         & UNETR      & 0.7368      & 0.8135     & \underline{0.7304}     & 0.6092 & 0.5769 & 0.5996 & 0.6104 & 0.6265 & 0.6007 & 0.5939  & \underline{0.6284} & 0.5890  \\
\hline
\multicolumn{1}{c}{\textbf{\textit{With Medical   SSL}}} &            &             &            &            &        &        &        &        &        &        &         &        &         \\
MG~\cite{zhou2021models}                   & CXR, CT        & UNET3D     & 0.7368      & \underline{0.9286}     & 0.4242     & 0.5756 & 0.5496 & 0.3653 & 0.6169 & 0.6980 & 0.6141 & 0.6121  & 0.6266 & 0.5892  \\
TransVW~\cite{haghighi2021transferable}              & CT              & UNET3D     & 0.7368      & 0.7222     & 0.4242     & 0.4958 & 0.6661 & 0.4450 & 0.6818 & 0.7228 & 0.6271 & \underline{0.6424}  & 0.5292 & 0.5003  \\
GVSL~\cite{he2023geometric}                   & CT              & UNET3D     & 0.7895      & 0.8516     & 0.7286     & 0.5966 & 0.6661 & 0.5959 & 0.6623 & \textbf{0.7309} & 0.6565 & 0.6242  & 0.5244 & 0.4701  \\
Swin-UNETR*~\cite{tang2022self}           & MRI             & Swin-UNETR & 0.7368      & 0.5032     & 0.4242     & 0.5462 & 0.5517 & 0.5461 & 0.6299 & 0.6437 & 0.5953 & 0.6303  & 0.4993 & 0.3866  \\
VoCo~\cite{wu2024voco}                  & MRI             & Swin-UNETR & 0.7368      & 0.5135     & 0.4242     & 0.5210 & 0.5740 & 0.5207 & 0.6558 & 0.6971 & 0.6413 & 0.5818  & 0.5626 & 0.5466  \\
DAE~\cite{valanarasu2023disruptive}                  & MRI             & Swin-UNETR & 0.7719      & 0.8151     & 0.7120     & 0.5294 & 0.5666 & 0.5294 & 0.6688 & 0.7129 & 0.6548 & 0.6061  & 0.5173 & 0.5548  \\
M$^3$AE~\cite{liu2023m3ae}                 & MRI             & UNET3D     & 0.7370      & 0.6984     & 0.5915     & 0.6008 & 0.6338 & 0.6003 & 0.6364 & 0.7049 & 0.6177 & 0.6061  & 0.5453 & 0.4769  \\
M$^3$AE~\cite{liu2023m3ae}               & MRI             & UniFormer  & \underline{0.7895}      & 0.8659     & 0.7159     & 0.6092 & 0.5352 & 0.5756 & 0.6169 & 0.6597 & 0.6028 & 0.5636  & 0.4682 & 0.4500  \\
\textbf{BrainMVP}         & MRI             & UNET3D     & 0.7895      & 0.7746     & 0.6621     & \underline{0.6555} & \underline{0.6669} & \underline{0.6421} & \underline{0.6818} & 0.7245 & \underline{0.6665} & \textbf{0.6970}  & 0.5817 & \textbf{0.6327}  \\
\textbf{BrainMVP}         & MRI             & UniFormer  & \textbf{0.8596}      & \textbf{0.9452}     & \textbf{0.8324}    & \textbf{0.6765} & \textbf{0.6964} & \textbf{0.6609} &\textbf{0.6883} & \underline{0.7249} &\textbf{0.6723} & 0.6182  & \textbf{0.6329} & \underline{0.5890} \\
\hline
\end{tabular}
}
\caption{Experimental results on four downstream \textbf{classification} datasets. We report the overall accuracy (ACC), area under the curve (AUC) and F1 score on each dataset. The best results are bolded and the second best results are underlined.}
\label{table:cls_res}
\end{table*}
\noindent{\textbf{Performance improvement on normal brain structure segmentation dataset.}} We utilize the MRBrainS13~\cite{mendrik2015mrbrains} dataset for the segmentation of normal brain structures to assess the efficacy of BrainMVP in scenarios with limited normal brain structure cases during pre-training. As detailed in Table~\ref{table:seg_res}, our BrainMVP achieves an average Dice Score of 80.27\%. In contrast, MG~\cite{zhou2021models}, employing multiple proxy tasks, attains 75.51\%, and VoCo~\cite{wu2024voco}, leveraging position prediction, achieves 76.37\%. Based on the UniFormer~\cite{li2023uniformer} architecture, BrainMVP surpasses all previous methods and demonstrates a notable 4.49\% average Dice Score improvement over training from scratch.
\begin{figure}[th]
    \centering
    \subfloat[BraTS2023-PED~\cite{kazerooni2023brain}]{
    \includegraphics[width=0.23\textwidth]{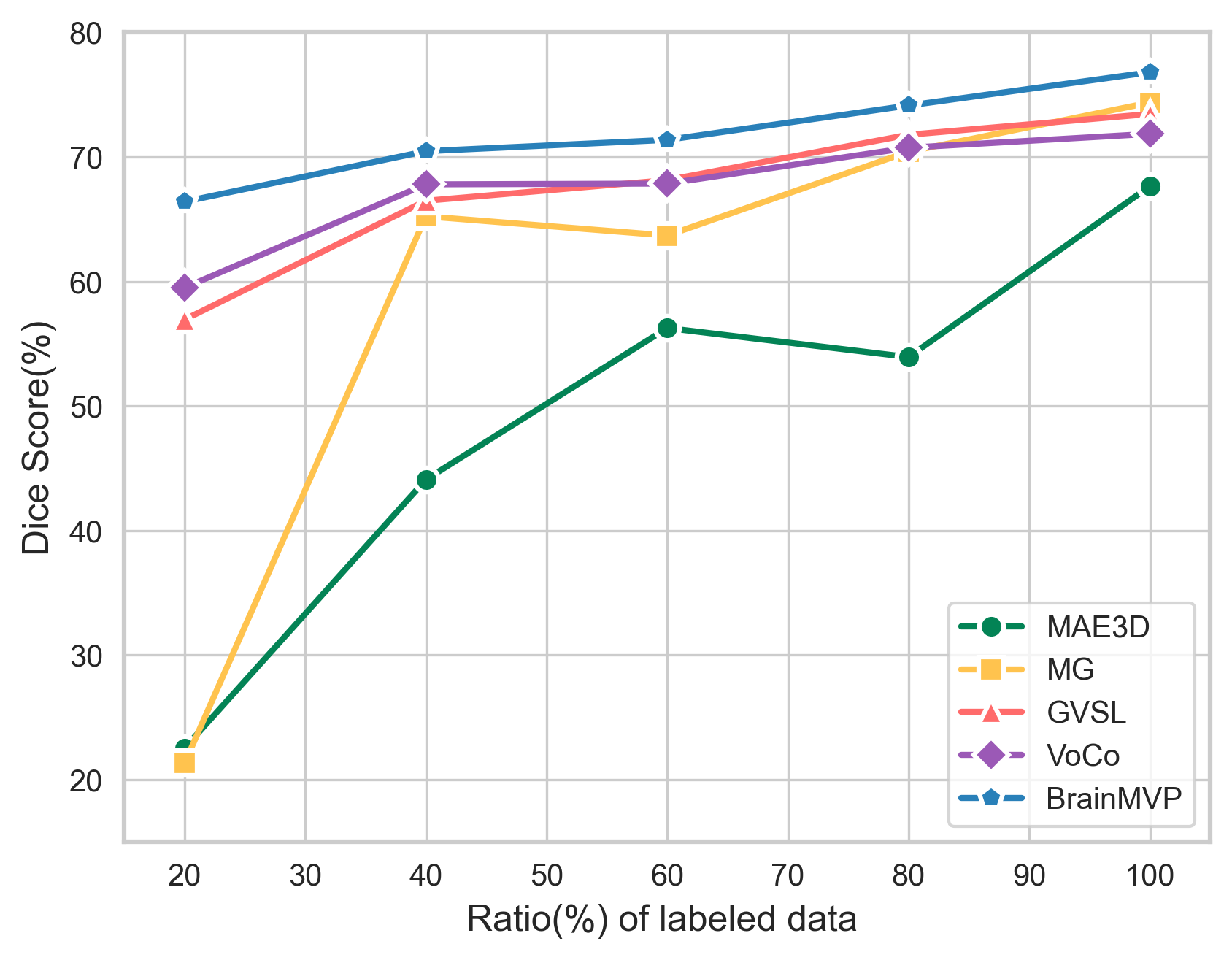}
    }
     \subfloat[BraTS-MET~\cite{moawad2023brain}]{
    \includegraphics[width=0.23\textwidth]{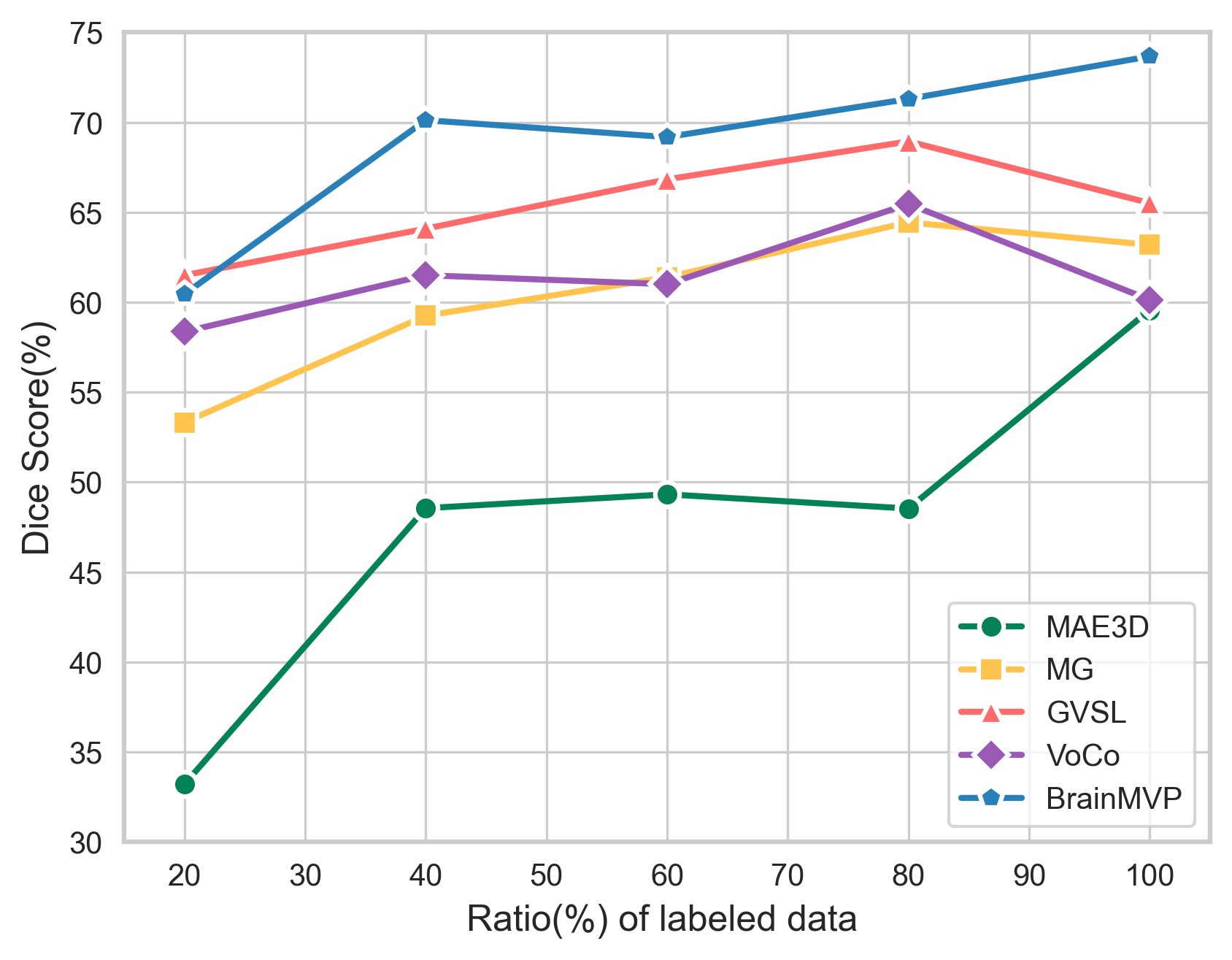}
    }
    
   \subfloat[ISLES22~\cite{hernandez2022isles}]{
    \includegraphics[width=0.23\textwidth]{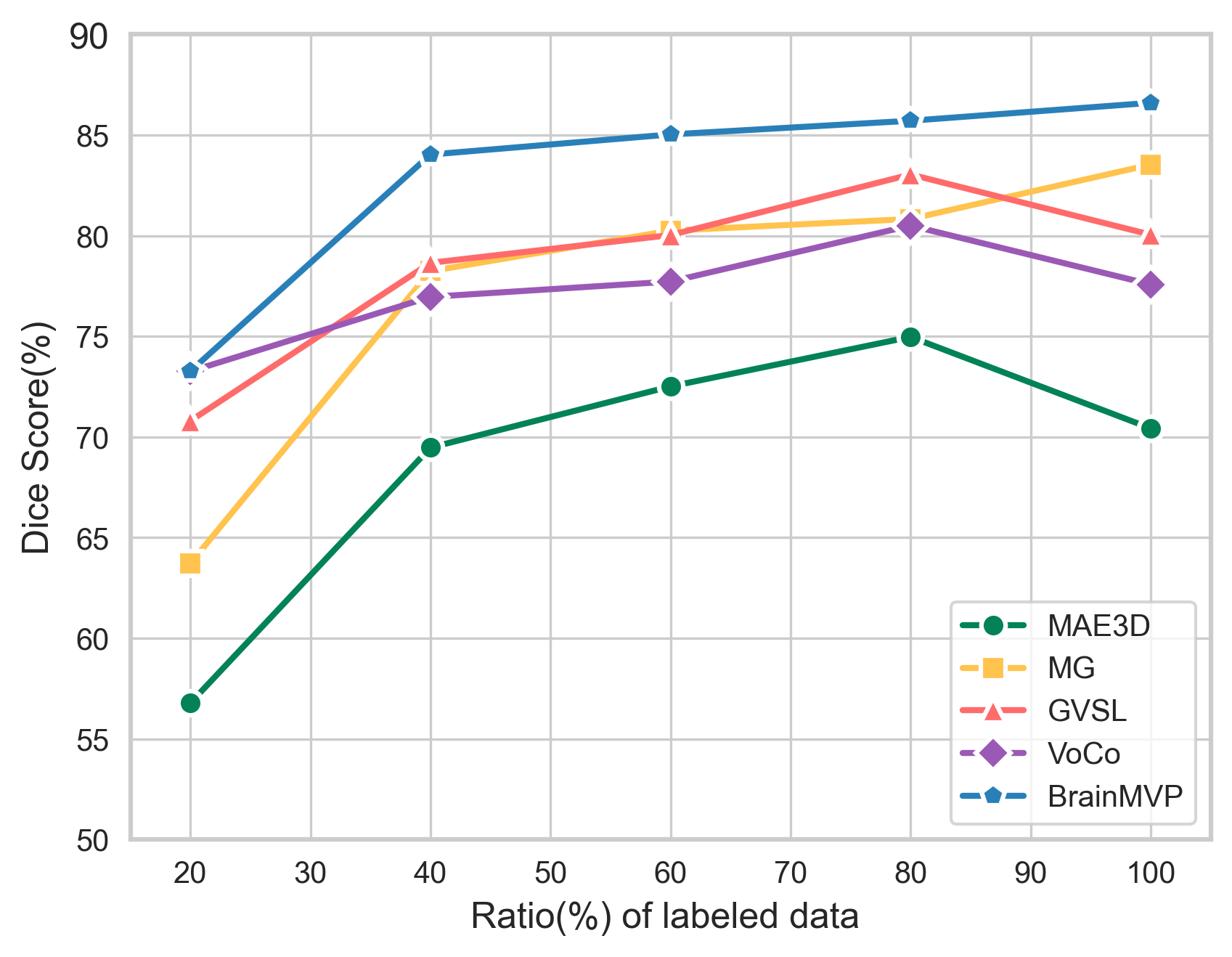}
    }
        \subfloat[VSseg~\cite{shapey2021segmentation}]{
    \includegraphics[width=0.23\textwidth]
    {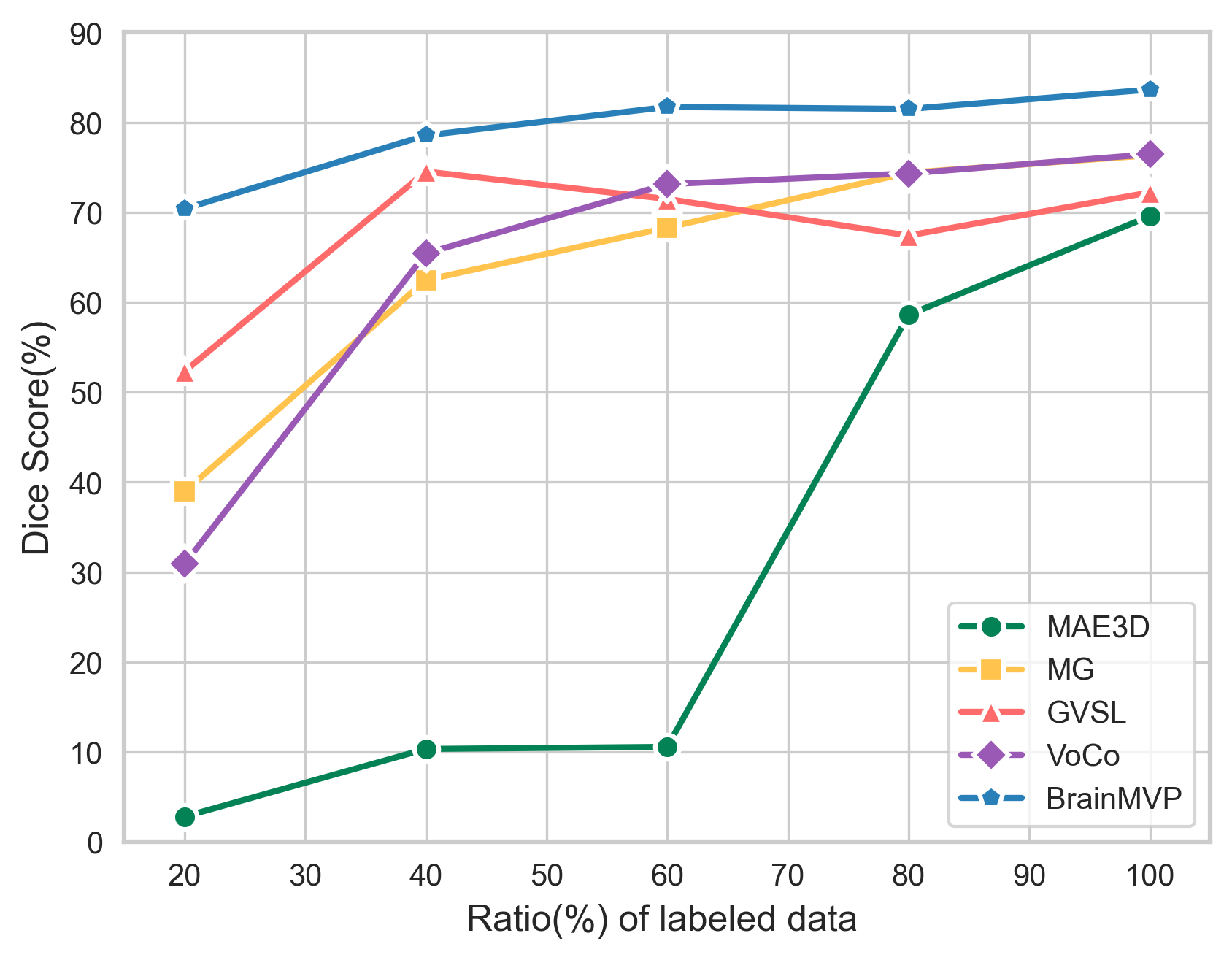}
    }
    
    \subfloat[UPENN-GBM~\cite{bakas2021multi}]{
    \includegraphics[width=0.23\textwidth]
    {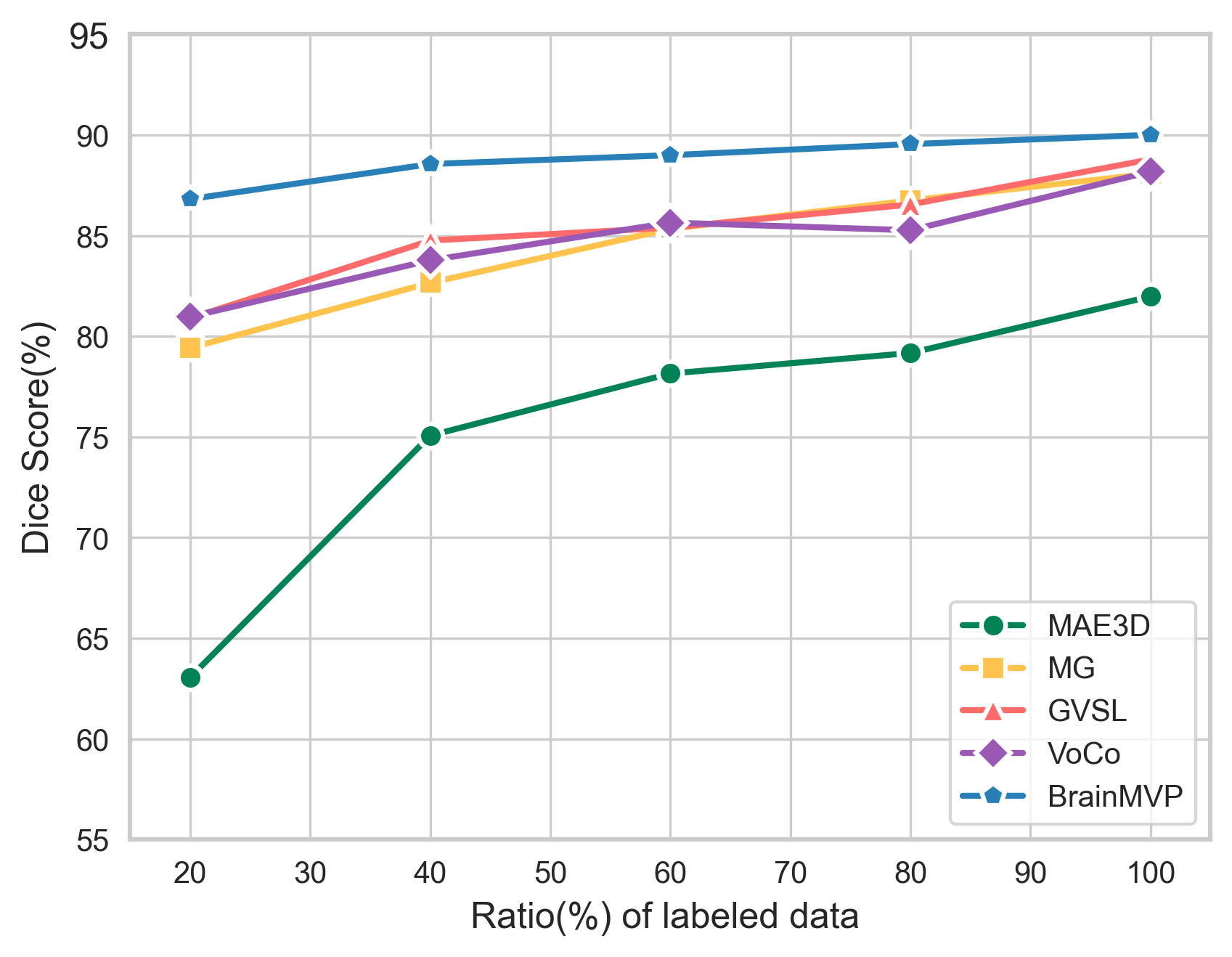}
    }
    \subfloat[BraTS2018~\cite{bakas2018identifying}]{
    \includegraphics[width=0.23\textwidth]{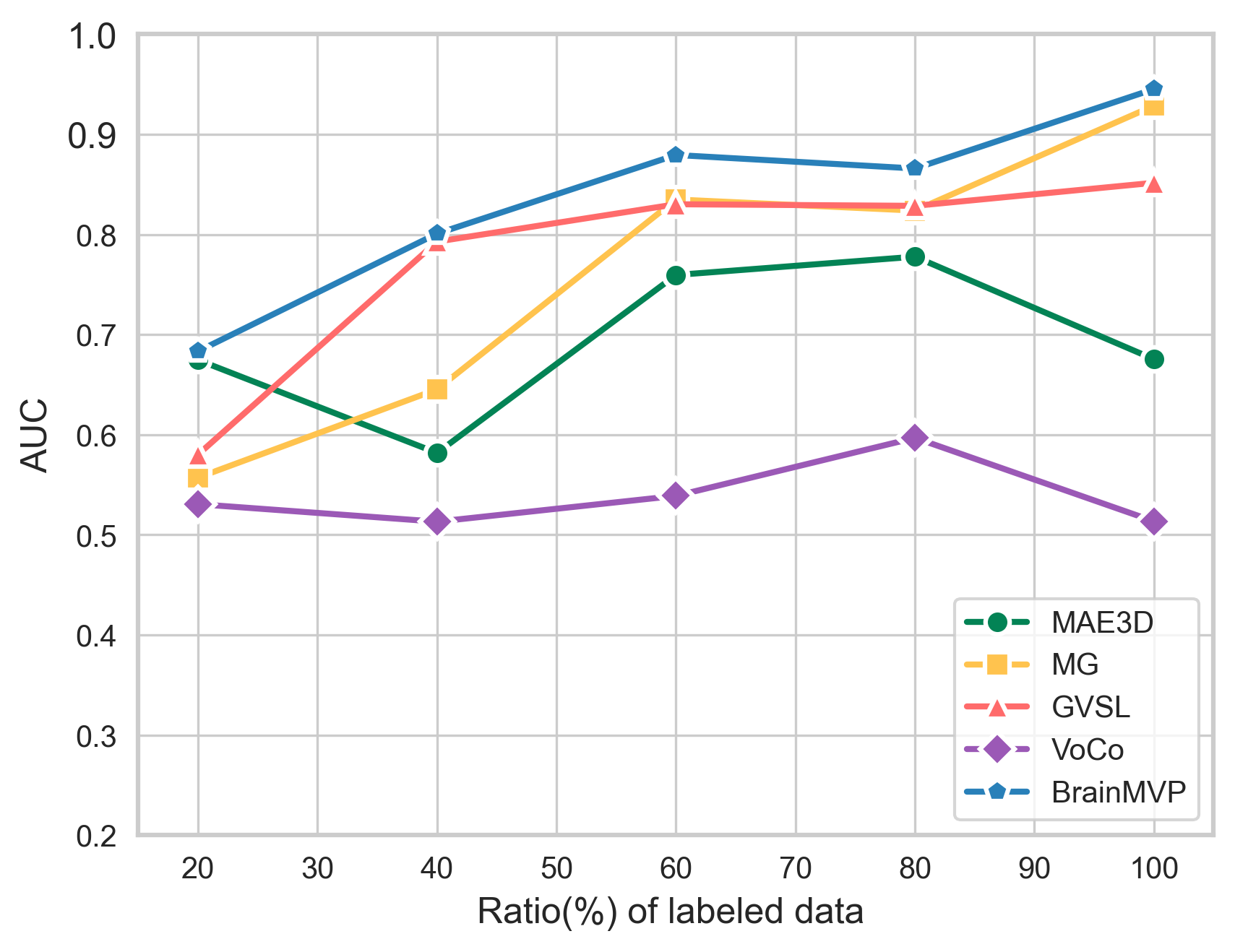}
    }
    \caption{Label efficiency results of the downstream segmentation and classification tasks. We report the mean Dice Score (\%) in segmentation and the area under the curve (AUC) in classification.}
    \label{fig:label-efficiency}
\end{figure}

\begin{table*}[h]
\centering
 \resizebox{0.8\textwidth}{!}{
\begin{tabular}{cccccccccc}
\hline
\multicolumn{3}{c}{Task}                         & BraTS2023-PED~\cite{kazerooni2023brain}             & \multicolumn{3}{c}{BraTS2018~\cite{bakas2018identifying}} & \multicolumn{3}{c}{ADNI~\cite{jack2008alzheimer}} \\
\hline
Recon.&Distill. & Contrast.&    Dice Score (\%) & ACC       & AUC       & F1       & ACC     & AUC    & F1    \\
\hline
\XSolidBrush            & \XSolidBrush          & \XSolidBrush                 & 72.52                  & 0.7762    & 0.7719    & 0.6994   & 0.5546           &0.6343        &0.5526       \\
 \CheckmarkBold             & \XSolidBrush          & \XSolidBrush                 & 75.16                  & 0.7895    & 0.8056    & 0.7286   & 0.6261        &0.6770        &0.5552       \\
\CheckmarkBold              & \CheckmarkBold          & \XSolidBrush                & 75.87                  & 0.8421    & 0.9032    & 0.8081   &  0.6261       &0.6835        &0.6187       \\
\CheckmarkBold              &\CheckmarkBold           & \CheckmarkBold                 & \textbf{76.80}                 &\textbf{0.8596} & \textbf{0.9452}    & \textbf{0.8324}   &       \textbf{0.6765}  &    \textbf{0.6964}   &\textbf{0.6609}     \\

\hline
\end{tabular}}
\caption{Ablation experimental results on BraTS2023-PED~\cite{kazerooni2023brain}, BraTS2018~\cite{bakas2018identifying} and ADNI~\cite{jack2008alzheimer} datasets. Recon.: cross-modal reconstruction; Distill.: Modality-wise data distillation; Contrast.: modality-aware contrastive learning. Note that cross-modal contrastive learning relies on the presence of both modules aforementioned.}
\label{table:ablation}
\end{table*}
\noindent{\textbf{Strong generalization performance on Unseen datasets.}} Given that our pre-training datasets primarily include normal brain structures and those afflicted with glioma, we aim to verify the generalization capabilities of BrainMVP on other types of diseases. To assess this, we evaluate our BrainMVP on three datasets: BraTS-MET~\cite{moawad2023brain}, ISLES22~\cite{hernandez2022isles}, and VSseg~\cite{shapey2021segmentation}. For the BraTS-MET~\cite{moawad2023brain} dataset focusing on brain metastasis subregion segmentation, as seen in Table~\ref{table:seg_res}, our BrainMVP achieves an average Dice Score of 73.67\%. Further, BrainMVP notably outperforms existing state-of-the-art methods in medical applications, including MG~\cite{zhou2021models} (63.19\%), and Swin-UNETR*~\cite{tang2022self} (63.23\%). In the context of the ISLES22~\cite{hernandez2022isles} ischemic stroke segmentation task, which involves abnormalities distinct from tumors targeted in pre-training, BrainMVP achieves substantial improvement compared to MG~\cite{zhou2021models} (86.60\% \textit{vs.} 83.53\%) and GVSL~\cite{he2023geometric} (86.60\% \textit{vs.} 80.05\%). For the VSseg~\cite{shapey2021segmentation} dataset focusing on vestibular schwannoma segmentation task, in previous methods, $\textrm{M}^{3}\textrm{AE}$~\cite{liu2023m3ae} achieves the best performance with 79.31\% Dice Score, while our BrainMVP outperforms all previous methods with 83.64\% Dice Score.\par
\noindent{\textbf{Classification Results}}.
We select four distinct classification tasks to assess the generalizability of BrainMVP across diverse domains. As illustrated in Table~\ref{table:cls_res}, on the BraTS2018~\cite{bakas2018identifying} dataset, our BrainMVP achieves an outstanding ACC of 0.8596, significantly surpassing the state-of-the-art M$^3$AE~\cite{liu2023m3ae} (0.7895), VoCo~\cite{wu2024voco} (0.7368), and GVSL~\cite{he2023geometric} (0.7895). BrainMVP also exhibits superior F1 score and AUC compared to prior SSL methods, highlighting its efficacy. Further experiments on ADNI~\cite{jack2008alzheimer}, ADHD-200~\cite{adhd2012adhd} and ABIDE-I~\cite{di2014autism} datasets show BrainMVP consistently outperforms state-of-the-art SSL methods. On ADHD-200~\cite{adhd2012adhd}, BrainMVP achieves an accuracy of 0.6883, surpassing the previous best one of 0.6818. On ABIDE-I~\cite{di2014autism}, BrainMVP improves the accuracy by 5.46\%, AUC by 0.45\%, and F1 score by 4.35\%.

\noindent{\textbf{High Label Efficiency}}:
Fig.~\ref{fig:label-efficiency} shows that BrainMVP consistently outperforms representative methods when fine-tuned on downstream tasks with varying labeled data ratios. As labeled data increases from 20\% to 40\%, BrainMVP significantly improves on multiple datasets: BraTS2023-PED~\cite{kazerooni2023brain} (Dice Score 66.41\% to 70.46\%), BraTS-MET~\cite{moawad2023brain} (60.45\% to 70.12\%), and ISLES22~\cite{hernandez2022isles} (73.27\% to 84.03\%). On BraTS2018~\cite{bakas2018identifying}, AUC rises from 0.6833 to 0.8008. Notably, with just 40\% labeled data, BrainMVP matches or exceeds fully labeled methods. With 20\% labeled data, BrainMVP achieves 66.41\% Dice Score on BraTS2023-PED~\cite{kazerooni2023brain}, 70.39\% on VSseg~\cite{shapey2021segmentation}, and 86.82\% on UPENN-GBM~\cite{bakas2021multi}, surpassing best-performing methods (59.50\%, 52.31\%, and 80.97\% respectively). This demonstrates BrainMVP's excellent efficiency, reducing annotation needs in clinical practice.

\subsection{Ablation Study}
We perform comprehensive ablation experiments on three key components of BrainMVP: cross-modal reconstruction, modality-wise data distillation, and modality-aware contrastive learning using representative BraTS2023-PED~\cite{kazerooni2023brain}, BraTS2018~\cite{bakas2018identifying}, and ADNI~\cite{jack2008alzheimer} datasets. The results are summarized in Table \ref{table:ablation}.

\noindent{\textbf{Cross-modal reconstruction}}:
As shown in Table \ref{table:ablation}, the inclusion of cross-modal reconstruction in pre-training leads to significant performance improvements. Specifically, on the BraTS2023-PED~\cite{kazerooni2023brain} dataset, the Dice Score increases from 72.52\% to 75.16\%, on the BraTS2018~\cite{bakas2018identifying} dataset, the AUC rises from 0.7719 to 0.8056, and on the ADNI~\cite{jack2008alzheimer} dataset, accuracy (ACC) improves from 0.5546 to 0.6261.
Notably, for the BraTS2023-PED~\cite{kazerooni2023brain} tumor subregion segmentation task, which requires detailed mpMRI information, the addition of cross-modal reconstruction significantly enhances performance. These results suggest that cross-modal reconstruction effectively captures modality associations, enabling more efficient multi-modal information fusion.\par
\noindent{\textbf{Modality-wise data distillation}}:
We next evaluate the effectiveness of the modality-wise data distillation module. As shown in Table \ref{table:ablation}, the AUC in the BraTS2018 tumor subtype classification task improves significantly, from 0.8056 to 0.9032. Consistent improvement can be seen in BraTS2023-PED~\cite{kazerooni2023brain} and ADNI~\cite{jack2008alzheimer} datasets. This suggests the distilled modality templates learned during pre-training enhance the diversity of downstream data, thereby improving BrainMVP's ability to generalize across tasks.\par
\noindent{\textbf{Modality-aware contrastive learning}}:
Finally, we investigate the impact of modality-aware contrastive learning. With its incorporation, BrainMVP's performance consistently improves across multiple datasets. On the BraTS2023-PED~\cite{kazerooni2023brain} dataset, the average Dice Score increases from 75.87\% to 76.80\%, and on the BraTS2018~\cite{bakas2018identifying} dataset for tumor subtype classification, the AUC rises from 0.9032 to 0.9452. For the ADNI~\cite{jack2008alzheimer} dataset, accuracy (ACC) improves from 0.6261 to 0.6765. Modality-aware contrastive learning, supported by cross-modal reconstruction and modality-wise data distillation, contributes to these gains. The combination of these components allows BrainMVP to achieve optimal results, demonstrating the effectiveness of the proposed pre-training framework.

%% file: sec/5_conclusion.tex
\vspace{-2pt}
\section{Conclusion}
In this paper, we propose BrainMVP, an efficient multi-modal vision pre-training method for multi-parametric brain MRI analysis. By exploiting structural similarities between MRI modalities, we design cross-modal reconstruction to capture modality correlations. To handle varying numbers of MRI modalities, we use single-channel images, enabling scalability. We also introduce modality-wise data distillation to learn condensed structural representations, and mix input modality images with condensed templates to link pre-training and downstream tasks. Additionally, modality-aware contrastive learning ensures semantic consistency and enhances the model’s discriminative ability. Extensive experiments on ten downstream datasets show that BrainMVP outperforms state-of-the-art methods and achieves strong generalizability. Our label efficiency experiment reveals that BrainMVP can match the performance of existing methods using only 40\% of labeled data, showcasing its potential for real-world clinical applications.

%% file: sec/6.tex
\vspace{-1pt}
\section*{Acknowledgments}
 This work was funded by the National Key R\&D Program of China (2022ZD0160700) and Shanghai Artificial Intelligence Laboratory.

%% file: sec/6_suppl.tex
\clearpage
\setcounter{page}{1}
\maketitlesupplementary
\appendix

\section{Distilled Modality Template for  Downstream Tasks}
\label{template_ds}
In this section, we will elaborate on how the distilled modality templates obtained from pre-training can be applied in downstream tasks.
\begin{figure*}
    \centering
    \includegraphics[width=0.7\linewidth]{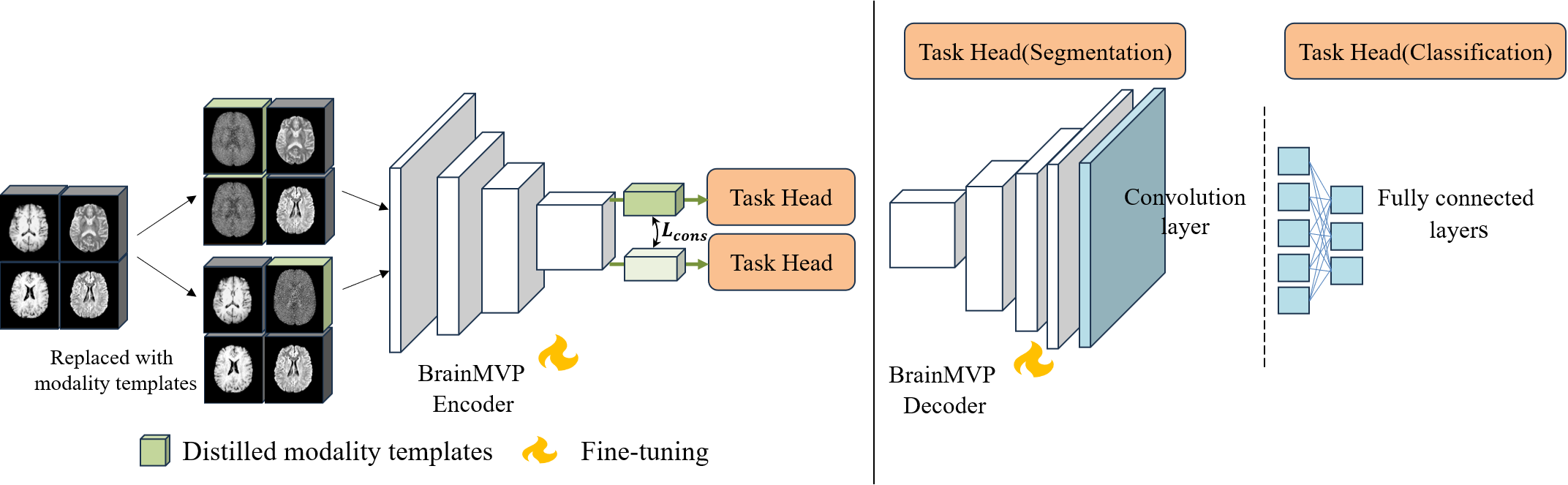}
    \caption{Modality-wise data distillation for \textbf{downstream tasks}. The input multi-modal MRI scans are randomly selected to replace a certain number of modalities with the corresponding modality templates. Then L2 norm is used to ensure feature consistency between the two replacement copies. Finally, the task head is replaced with corresponding modules based on the task type.}
    \label{fig:data distillation}
\end{figure*}
As shown in Fig.~\ref{fig:data distillation}, in the downstream fine-tuning stage, the distilled modality templates are frozen. Let $\mathcal{D}_{ds}=\{(X_{i},Y_{i})\}_{i=1}^M$ denote the downstream dataset, where $M$ represents the number of annotated samples. $X_{i}$ is the multi-modal MRI input volume, and $Y_{i}$ represents the corresponding label, which can be a segmentation map for segmentation tasks or a one-hot vector for classification tasks. Specifically, we randomly select $m$ and $n$ modalities in $X_{i}$ and replace them with the corresponding modalities from $\{T_{m}\}_{m=1}^S$, obtaining two augmented copies $X_{i}'$ and $X_{i}''$. The encoded features of these two copies are $\mathcal F_{enc}({X_{i}'})$ and $\mathcal F_{enc}({X_{i}''})$, respectively. Since the two embeddings are representations of the same sample with different numbers of replaced modalities, we use the L2 norm to maintain semantic consistency in the feature space.
\begin{align}
    \mathcal L_{cons} = || \mathcal F_{enc}({X_i'})-\mathcal F_{enc}({X_i''})||_2
\end{align}\par
Subsequently, the features of the two copies are decoded to the output space to calculate supervision loss with the ground-truth annotations. The overall fine-tuning loss is: 

\begin{equation}
\begin{aligned}
    \mathcal{L}_{FT} = &\frac{1}{|\mathcal{B}|} \sum_{i=1}^{|\mathcal{B}|} \left( \mathcal{L}_{sl}(\mathcal{F}(X_i'), Y_i) + \mathcal{L}_{sl}(\mathcal{F}(X_i''), Y_i) \right. \\
    &\left. + \lambda_{cons} * \mathcal{L}_{cons} \right)
\end{aligned}
\end{equation}

where $\lambda_{cons}$ is the weight of the consistency loss $\mathcal L_{cons}$ term and $\mathcal L_{sl}$ is the supervision loss used in segmentation or classification tasks, e.g., Dice Loss in segmentation or Cross-Entropy Loss in classification. $|\mathcal B|$ represents number of cases in a batch.\par
For the uni-modal input scenario, instead of replacing the selected modalities with distilled modality templates, we perform a partially masking strategy like Algorithm~\ref{alg:alg1} where $X_{i}$ is replaced with the corresponding distilled modality template. Then we randomly mask the uni-modal input volume twice to obtain two augmented copies of $X_{i}$, and the remaining procedures are the same as the aforementioned multi-modal scenario.
\par

\section{Pre-processing}
\label{sec:imp_detail}
\subsection{Pre-training}
During pre-training, data pre-processing is performed sequentially in Python based on MONAI 1.3.0\footnote{\url{https://monai.io/}} library. The orientation of the mpMRI scan is first unified to the RAS axcodes and co-registered to the same anatomical template. Subsequently, each MRI scan is resampled to an isotropic voxel spacing of $1.0mm\times1.0mm\times1.0 mm$ using bilinear interpolation, and skull-stripping is performed as well. We linearly clip the pixel values between the 1st and 99th percentiles and re-scale them to [0, 1]. The images are then cropped into $96\times96\times96$ voxel patches centered 
on either foreground or background areas, to ensure that the modality-wise data distillation is learned sufficiently. We do not apply any other data augmentation techniques.\par

\subsection{Segmentation}
The input mpMRI scan is first reoriented to the RAS coordinate system, then the image spacing is adjusted to a uniform $1.0mm\times1.0mm\times1.0mm$ ( for the ISLES22~\cite{hernandez2022isles} dataset it’s $1.5mm\times1.5mm\times1.5mm$ ) using bilinear interpolation. Subsequently, the pixel grayscale values of the input mpMRI scan are normalized from the 5th to the 95th percentile, with each channel being adjusted to a range between 0 and 1. After cropping the foreground area of the image, we randomly crop a fixed area of $96\times96\times96$. To avoid over-segmentation, we allow the sampling center to be in the background area. Then, random mirror flipping along three axes with a probability of 0.5, random intensity offset with 0.1 offset, random intensity scaling with probability 1.0 in a scale factor of 0.1 are performed for data augmentation. For network training, we employ the AdamW optimizer~\cite{loshchilov2017decoupled} with an initial learning rate of 3e-4, incorporating cosine learning rate decay. Weight decay is set to 1e-3 for UNETR~\cite{hatamizadeh2022unetr}-based models, 1e-4 for UniFormer~\cite{li2023uniformer} and Swin-UNETR~\cite{hatamizadeh2021swin}-based models, and 1e-5 for UNET3D~\cite{ronneberger2015u}-based models. We train the network with a batch size of 3 for 500 epochs, and $\lambda_{cons}$is set to 0.1.\par

\subsection{Classification} 
The data augmentation part is different from segmentation in that we resize the input image to a fixed size of $128\times128\times64$ after normalizing it to fit the training of the comparison methods. Subsequently, we randomly crop a fixed region of $96\times96\times64$ and then perform the same random data augmentation as segmentation. In the inference stage, we crop an area of $96\times96\times64$ at the center of the input image. we set the batch size to 64 considering gradient accumulation and train all networks for 200 epochs. The remaining hyper-parameters are the same as those used for segmentation.

\section{Dataset Details}
\subsection{Pre-training datasets}
\label{pre_dataset}
\noindent{\textbf{BraTS2021~\cite{baid2021rsna}}}: This dataset comprises 1,470 cases publicly available multi-sequence MRI scans, encompassing four paired modalities: T1, T1CE, T2, and FLAIR. All images have been registered and resampled to $1.0mm\times1.0mm\times1.0mm$. We only utilize the image data without incorporating the segmentation annotations.

\noindent{\textbf{BraTS2023-SSA~\cite{adewole2023brain}}} and \textbf{BraTS2023-MEN}~\cite{labella2023asnr}: These datasets are two of the five segmentation sub-tasks in BraTS2023 with 75 cases and 1,141 cases mpMRI, respectively. The former dataset focuses on the segmentation of brain gliomas in patients from sub-Saharan Africa, while the latter is dedicated to adult meningioma segmentation. Note that the modality type is identical to \textbf{BraTS2021~\cite{baid2021rsna}}, albeit involving a different type of brain tumor.

\noindent{\textbf{UCSF-PDGM~\cite{calabrese2022university}}}: This dataset comprises 501 cases with various mpMRI data, from which we select six modalities--T1, T1CE, T2, FLAIR, DWI, and ADC for corresponding downstream applications.

\noindent{\textbf{IXI}}:\footnote{https://brain-development.org/ixi-dataset/} This dataset includes 600 MR images from normal, healthy subjects with T1, T2, PD, MRA and DTI images. We select 568 cases that include all four modalities: T1, T2, PD, and MRA for pre-training and this dataset serves as a supplement to the pre-training brain dataset, specifically for normal brain cases.

\subsection{Downstream datasets
}
\label{ds_details}
We conduct a comprehensive evaluation using ten downstream datasets encompassing segmentation and classification tasks. The details are as follows:

\noindent{\textbf{Segmentation}}: (1) BraTS2023-PED~\cite{kazerooni2023brain}: This dataset comprises 99 publicly annotated pediatric brain glioma multi-sequence MRI scans. The annotations include Non-Enhancing Core (NEC), Edema, and Enhancing Tumor (ET). (2) BraTS2023-MET~\cite{moawad2023brain}: Similarlly, this dataset focuses on brain metastasis sub-region segmentation from multi-sequence MRI. It contains 238 publicly available imaging cases with four modalities: T1, T1CE, T2W, and FLAIR. (3) ISLES22~\cite{hernandez2022isles}: This dataset aims to segment acute to subacute ischemic stroke lesions from multi-sequence MR images (including FLAIR, DWI, and ADC). We collected 238 publicly annotated cases. (4) MRBrainS13~\cite{mendrik2015mrbrains}: This dataset targets brain structure segmentation from 20 cases with three sequences: T1, T1CE, and FLAIR MR images. The segmentation targets include Cerebrospinal Fluid (CF), Gray Matter (GM), and White Matter (WM). (5) UPENN-GBM~\cite{bakas2021multi}: We collected 127 publicly annotated multi-sequence MR images from de novo Glioblastoma (GBM) patients, similarly focusing on segmenting three tumor subregions. (6) VSseg~\cite{shapey2021segmentation}: This dataset includes 242 cases of multi-sequence MRI data from patients with vestibular schwannoma, aiming to segment the vestibular schwannoma region.

\noindent{\textbf{Classification}}: (1) BraTS2018~\cite{bakas2018identifying}: This dataset includes a tumor subtype classification task, aiming to determine the severity grade of brain tumors from four MR modalities, labeled as HGG (High-Grade Glioma) or LGG (Low-Grade Glioma). (2) ADNI~\cite{jack2008alzheimer}: This dataset represents late-life brain disorders through Alzheimer's Disease (AD) cases. Given the importance of early diagnosis, we analyze the most recent neuroimaging scans and demographic data from 1348 subjects, labeled as mild cognitive impairment (MCI) or normal control (NC). (3) ADHD-200~\cite{adhd2012adhd} and (4) ABIDE-I~\cite{di2014autism}: These two datasets are utilized for early-life brain disorder studies. For ADHD-200~\cite{adhd2012adhd}, T1-weighted MRI scans and demographic information (age and gender) are collected from 767 subjects, including 279 ADHD patients and 488 controls. ABIDE-I~\cite{di2014autism} comprises neuroimaging data from 819 subjects (327 with autism spectrum disorder and 492 typically developing controls) with matching imaging modalities.

The aforementioned datasets, except for MRBrainS13~\cite{mendrik2015mrbrains}, are randomly partitioned into training, validation, and test sets with a ratio of 6:1:3. For MRBrainS13~\cite{mendrik2015mrbrains}, 5 cases are used for training and the remaining 15 cases for testing. It's worth noting that the data splits for ADNI~\cite{jack2008alzheimer}, ADHD-200~\cite{adhd2012adhd}, and ABIDE-I~\cite{di2014autism} datasets are performed at the patient/case level, ensuring that scans from the same subject will not appear across different sets.

\begin{algorithm}[H]
\caption{Pixel-level cross-modal masking.}\label{alg:alg1}
\begin{algorithmic}
\STATE Sample randomly $X_{im}$ from $X_{i}$
\STATE Sample randomly $X_{in}(n\neq m)$ from $X_{i}$
\STATE $p_{total} \leftarrow H \times W \times D$ 
\STATE $p_{mask} \leftarrow 0$ 
\WHILE{$p_{mask} < p_{total}\times p^*$}
    \STATE  Select randomly $(x, y, z)$ in $X_{im}$
    \STATE Mask an area of size $r \times r \times r$ centered at $(x, y, z)$
    \STATE Fill with corresponding  data from $X_{in}$
    \STATE $p_{mask}\leftarrow p_{mask} \bigoplus r\times r\times r$
\ENDWHILE
\RETURN modified ${X}_{im}$
\end{algorithmic}
\label{alg1}
\end{algorithm}

\section{HD95 Results and Visualization}
In Table~\ref{table:hd95_1} and Table~\ref{table:hd95_2}, we report the HD95 metric results of the pre-trained model on segmentation and classification tasks, respectively. These experimental results indicate that BrainMVP consistently exhibits smaller structural errors.\par
\begin{figure*}[t]
    \centering
    \includegraphics[width=\linewidth]{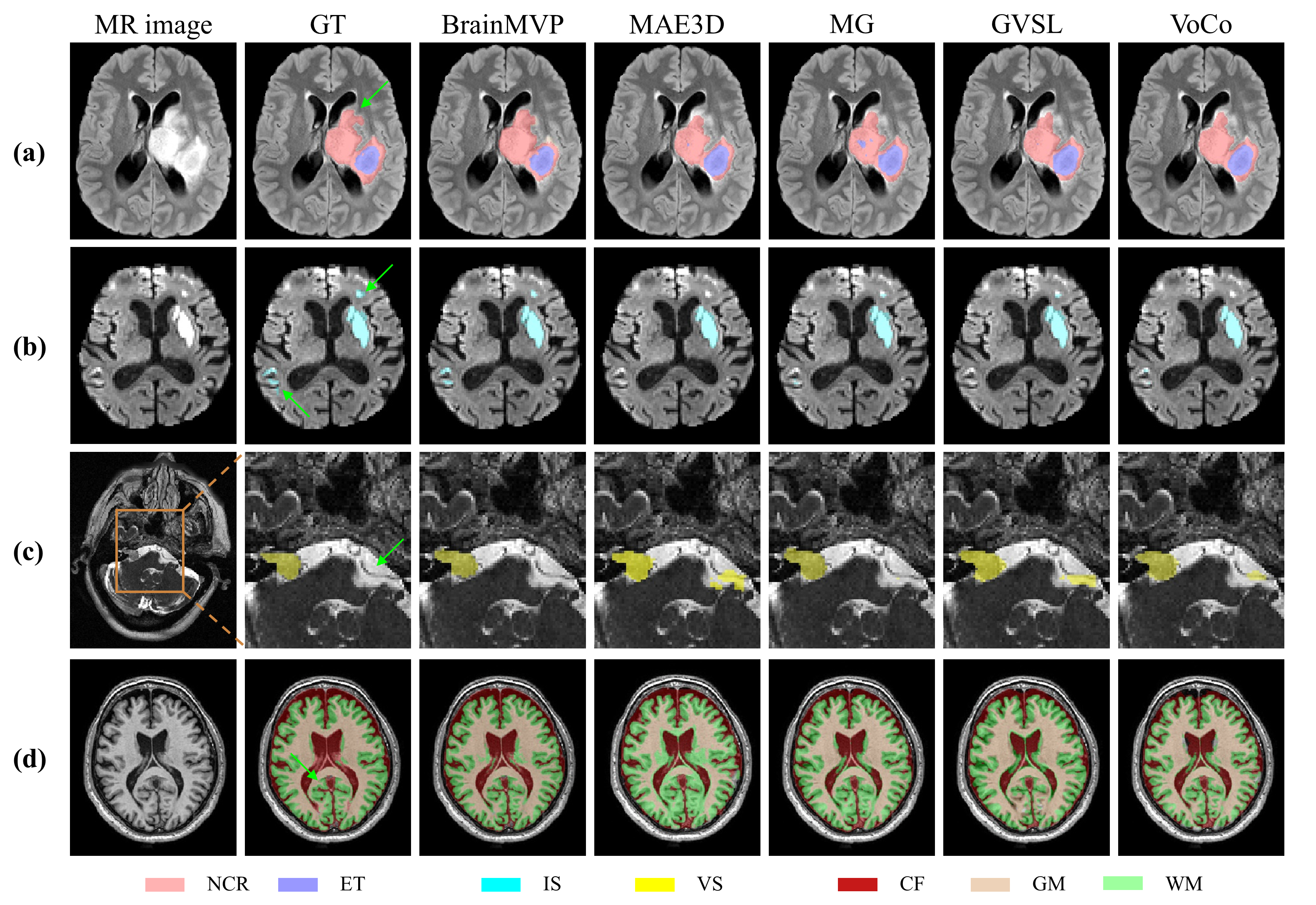}
    \caption{Visualization results of segmentation tasks. (a) BraTS2023-PED~\cite{kazerooni2023brain}: pediatric tumor subregion segmentation. NCR: necrotic tumor core; ET: enhancing tumor. (b) ISLES22~\cite{hernandez2022isles}: Ischemic Stroke lesion (IS) segmentation. (c) VSseg~\cite{shapey2021segmentation}: Vestibular schwannoma (VS) segmentation. (d) MRBrainS13~\cite{mendrik2015mrbrains}: brain structure segmentation. CF: Cerebrospinal Fluid; GM: Gray matter; WM: White matter. GT: ground truth. The green arrows highlight the regions where BrainMVP demonstrates superior performance over other methods.}
    \label{fig:visualization}
\end{figure*}

To facilitate qualitative comparison, we visualize the results obtained from MAE3D~\cite{he2022masked,chen2023masked}, MG~\cite{zhou2021models}, GVSL~\cite{he2023geometric}, VoCo~\cite{wu2024voco}, and BrainMVP on four datasets. The visualizations are shown in Fig.~\ref{fig:visualization}. The visualization results indicate that our BrainMVP segmentation results are most consistent with the ground truth (GT), significantly mitigating the issues of under-segmentation and over-segmentation. As shown in Fig.~\ref{fig:visualization} (a) for the NCR region boundary, BrainMVP demonstrates more accurate identification, while other methods exhibit substantial under-segmentation.

\begin{table}[tb]
\caption{Experimental results on datasets BraTS2023-PED~\cite{kazerooni2023brain}, BraTS2023-MET~\cite{moawad2023brain} and ISLES22~\cite{hernandez2022isles}. We report the mean HD95 ($\downarrow$) on each dataset.}
  \setlength{\tabcolsep}{2.5pt} %

 \resizebox{0.5\textwidth}{!}{
\begin{tabular}{cccccccccccc}
\hline
\textbf{Method }              &\textbf{ Modality}        & \textbf{Network}    & \multicolumn{4}{c}{\textbf{BraTS2023-PED~\cite{kazerooni2023brain}}}                                                                      & \multicolumn{4}{c}{\textbf{BraTS2023-MET~\cite{moawad2023brain}}}                                                                      & \multicolumn{1}{l}{\textbf{ISLES22~\cite{hernandez2022isles}}}                                                \\
\hline
                     &                 &            & ET & TC& WT & AVG& ET &TC & WT & AVG & IS    \\
                     \hline
\multicolumn{1}{c}{\textbf{\textit{From Scratch}}}       &            & 
\multicolumn{1}{l}{}   & \multicolumn{1}{l}{}   & \multicolumn{1}{l}{}   & \multicolumn{1}{l}{}    & \multicolumn{1}{l}{}   & \multicolumn{1}{l}{}   & \multicolumn{1}{l}{}     \\

UNETR~\cite{hatamizadeh2022unetr}                & -               & -          & 25.06                  & 39.07                  & 39.14                  & 34.43                   & 44.11                  & 45.22                  & 43.36                  & 44.23                   & 15.48                    \\
UNET3D~\cite{ronneberger2015u}               & -               & -          & 22.48                  & 34.02                  & 33.07                  & 29.86                   & 45.68                  & 46.85                  & 39.93                  & 44.15                   & 4.43                      \\
UniFormer~\cite{li2023uniformer}            & -               & -          & \multicolumn{1}{l}{11.55}   & \multicolumn{1}{l}{16.71}   & \multicolumn{1}{l}{16.14}   & \multicolumn{1}{l}{14.80}    & \multicolumn{1}{l}{25.90}   & \multicolumn{1}{l}{28.16}   & \multicolumn{1}{l}{19.97}   &       24.68                  &    4.13       \\
Swin-UNETR~\cite{hatamizadeh2021swin}           & -               & -          & 17.37                  & 22.56                  & 21.03                  & 20.32                   & 28.68                  & 31.03                  & 24.26                  & 27.99                   & 11.31                       \\
\hline
\multicolumn{2}{l}{\textbf{\textit{With General   SSL}}} &            & \multicolumn{1}{l}{}   & \multicolumn{1}{l}{}   & \multicolumn{1}{l}{}   & \multicolumn{1}{l}{}    & \multicolumn{1}{l}{}   & \multicolumn{1}{l}{}   & \multicolumn{1}{l}{}   &                         & \multicolumn{1}{l}{}             \\
MAE3D~\cite{he2022masked,chen2023masked}               & Natural         & UNETR      & 25.37                  & 38.43                  & 37.92                  & 33.90                   & 36.89                  & 36.57                   & 38.38                  & 37.28                   & 15.20                       \\
SimMIM~\cite{xie2022simmim}               & Natural         & UNETR      & 24.70                  & 31.61                  & 32.52                  & 29.61                   & 39.37                  & 41.26                  & 40.06                  & 40.23                   & 17.14                      \\
MoCoV3~\cite{chen2021empirical}              & Natural         & UNETR      & 20.60                  & 31.88                  & 32.12                  & 28.20                   & 41.88                  & 43.17                  & 41.92                  & 42.32                   & 15.04                 \\
\hline
\multicolumn{2}{l}{\textbf{\textit{With Medical   SSL}}} &            & \multicolumn{1}{l}{}   & \multicolumn{1}{l}{}   & \multicolumn{1}{l}{}   & \multicolumn{1}{l}{}    & \multicolumn{1}{l}{}   & \multicolumn{1}{l}{}   & \multicolumn{1}{l}{}   &                         & \multicolumn{1}{l}{}          \\
MG~\cite{zhou2021models}                   & CXR, CT        & UNET3D     & 19.71                  & 15.72                  & 17.65                  & 17.69                   & 46.39                  & 48.33                  & 42.02                  & 45.58                   & 3.68                     \\
TransVW~\cite{haghighi2021transferable}             & CT              & UNET3D     & 18.36                  & 25.42                  & 24.67                  & 22.82                   & 47.85                  & 48.06                  & 39.41                  & 45.11                   & 7.93                    \\
GVSL~\cite{he2023geometric}                 & CT              & UNET3D     & 17.45                  & 15.33                  & 16.00                  & 16.26                   & 37.33                  & 38.05                  & 30.61                  & 35.33                   & 9.35             \\
Swin-UNETR*~\cite{tang2022self}          & MRI             & Swin-UNETR & 18.65                  & 17.44                  & 17.64                  & 17.91                   & 40.57                  & 41.54                  & 33.93                  & 38.68                   & 8.09             \\
VoCo~\cite{wu2024voco}                 & MRI             & Swin-UNETR & 18.98                  & 17.21                  & 17.16                  & 17.78                   & 38.52                  & 39.79                  & 34.73                  & 37.68                   & 12.22                    \\
DAE~\cite{valanarasu2023disruptive}                  & MRI             & Swin-UNETR & 19.33                  & 21.41                  & 21.71                  & 20.82                   & 37.63                  & 37.37                  & 38.74                  & 37.91                   & 12.50                     \\
M$^3$AE~\cite{liu2023m3ae}                 & MRI             & UNET3D     & 13.48   & 11.91   &10.88  & 12.09   & 22.40  &23.87 & 18.96 &         21.74                & 4.58     \\
M$^3$AE~\cite{liu2023m3ae}                 & MRI             & UniFormer  & 16.19   & 15.95  & 19.78  & 17.31  & 25.89 & 28.37  & 24.35  &    26.21                     & 2.64         \\
\textbf{BrainMVP}         & MRI             & UNET3D     & 15.93                  & 7.24                  & 9.81                  & 10.99                   & 20.37                  & 22.50                  & 18.34                  & 20.40                   & 5.85                     \\
\textbf{BrainMVP}         & MRI             & UniFormer  & 13.93  & 7.88   & 14.56   & 12.12    & 22.60   & 25.88   & 19.83 & 22.77  & 2.69        \\
\hline
\end{tabular}}
\par
\vspace{0.5em}
{\fontsize{7pt}{20pt}\selectfont \quad CXR: Chest X-Ray; ET: enhancing tumor; TC: tumor core; WT: whole tumor; AVG: average; CF: Cerebrospinal Fluid; GM: Gray matter; WM: White matter; IS: Ischemic Stroke.}
\label{table:hd95_1}
\end{table}

\begin{table}[tb]
\caption{Experimental results on datasets MRBrainS13~\cite{mendrik2015mrbrains}, VSseg~\cite{shapey2021segmentation} and UPENN-GBM~\cite{bakas2021multi}. We report the mean HD95 ($\downarrow$) on each dataset.}
  \setlength{\tabcolsep}{2.5pt} %

 \resizebox{0.5\textwidth}{!}{
\begin{tabular}{cccccccccccc}
\hline
\textbf{Method}           &\textbf{ Modality}        & \textbf{Network}     &\multicolumn{4}{c}{\textbf{MRBrainS13~\cite{mendrik2015mrbrains}}}& \multicolumn{1}{l}{\textbf{VSseg~\cite{shapey2021segmentation}}} & \multicolumn{4}{c}{\textbf{UPENN-GBM~\cite{bakas2021multi}}}\\
\hline
                     &                 &              & CF& GM &WM& AVG &VS  & ET & TC & WT & AVG   \\
                     \hline
\multicolumn{1}{c}{\textbf{\textit{From Scratch}}}       &            &   &  &  &    &     &      &   &          & &    \\
UNETR~\cite{hatamizadeh2022unetr}                & -               & -                            & 4.16                                    & 3.46                            & 5.04                               & 4.22                    & 24.54                     & 16.97                  & 24.80                  & 31.00                  & 24.26 \\
UNET3D~\cite{ronneberger2015u}               & -               & -                            & 3.24                                    & 2.91                            & 3.70                               & 3.28                    & 34.36                     & 5.30                   & 9.34                   & 13.31                  & 9.32  \\
UniFormer~\cite{li2023uniformer}            & -               & -             &           2.38         &    2.43         &    4.04       &  2.95  &   5.68   &  4.46  &  6.97  & 11.32&   7.58    \\
Swin-UNETR~\cite{hatamizadeh2021swin}           & -               & -                           & 3.38                                    & 2.65                            & 4.00                               & 3.34                    & 14.12                     & 1.86                   & 7.22                   & 9.15                   & 6.08  \\
\hline
\multicolumn{2}{l}{\textbf{\textit{With General   SSL}}} &                    & \multicolumn{1}{l}{}                    & \multicolumn{1}{l}{}            & \multicolumn{1}{l}{}               & \multicolumn{1}{l}{}    & \multicolumn{1}{l}{}      & \multicolumn{1}{l}{}   & \multicolumn{1}{l}{}   & \multicolumn{1}{l}{}   &       \\
MAE3D~\cite{he2022masked,chen2023masked}               & Natural         & UNETR                        & 3.69                                    & 2.62                            & 3.59                               & 3.30                    & 24.17                     & 15.41                  & 20.10                  & 35.71                  & 23.74 \\
SimMIM~\cite{xie2022simmim}               & Natural         & UNETR                     & 3.84                                    & 2.67                            & 3.55                               & 3.35                    & 26.82                     & 17.23                  & 20.71                  & 32.11                  & 23.35 \\
MoCoV3~\cite{chen2021empirical}              & Natural         & UNETR            & 3.84                                    & 2.99                            & 4.74                               & 3.86                    & 21.35                     & 17.08                  & 19.83                  & 34.35                  & 23.75 \\
\hline
\multicolumn{2}{l}{\textbf{\textit{With Medical   SSL}}} &               & \multicolumn{1}{l}{}                    & \multicolumn{1}{l}{}            & \multicolumn{1}{l}{}               & \multicolumn{1}{l}{}    & \multicolumn{1}{l}{}      & \multicolumn{1}{l}{}   & \multicolumn{1}{l}{}   & \multicolumn{1}{l}{}   &       \\
MG~\cite{zhou2021models}                   & CXR, CT        & UNET3D                       & 3.47                                    & 9.43                            & 12.67                              & 8.52                    & 14.87                     & 2.27                   & 4.29                   & 12.67                  & 6.41  \\
TransVW~\cite{haghighi2021transferable}             & CT              & UNET3D                        & 3.81                                    & 3.45                            & 2.93                               & 3.40                    & 16.83                     & 3.36                   & 5.73                   & 12.95                  & 7.35  \\
GVSL~\cite{he2023geometric}                 & CT              & UNET3D                     & 3.73                                    & 3.44                            & 3.28                               & 3.48                    & 11.58                     & 2.23                   & 3.71                   & 9.17                   & 5.03  \\
Swin-UNETR*~\cite{tang2022self}          & MRI             & Swin-UNETR                           & 3.33                                    & 2.26                            & 2.33                               & 2.64                    & 20.73                     & 2.44                   & 4.07                   & 9.79                   & 5.43  \\
VoCo~\cite{wu2024voco}                 & MRI             & Swin-UNETR                   & 3.14                                    & 3.88                            & 7.87                               & 4.96                    & 13.26                     & 28.50                  & 43.05                  & 31.51                  & 34.35 \\
DAE~\cite{valanarasu2023disruptive}                  & MRI             & Swin-UNETR                       & 3.07                                    & 2.27                            & 3.36                               & 2.90                    & 19.84                     & 2.24                   & 3.90                   & 9.56                   & 5.23  \\
M$^3$AE~\cite{liu2023m3ae}                 & MRI             & UNET3D          & 3.69                  & 3.88          & 3.01            & 3.53   & 9.20    & 1.85   & 4.65   &8.24  &    4.91   \\
M$^3$AE~\cite{liu2023m3ae}                 & MRI             & UniFormer        & 1.89                & 2.92        & 4.53         & 3.11   & 9.16      &     4.75                   &     6.54                   &        9.93                &    7.07   \\
\textbf{BrainMVP}         & MRI             & UNET3D                         & 3.71                 & 4.92           &3.84              & 4.14   & 16.41      & 2.35                   & 4.60                   & 9.13                  & 5.36  \\
\textbf{BrainMVP}         & MRI             & UniFormer       & 1.53                    & 5.60          & 7.02            & 4.72  & 6.00      & 1.48  & 6.66   & 10.59  &  6.24   \\
\hline
\end{tabular}}
\par
\vspace{0.5em}
{\fontsize{7pt}{20pt}\selectfont \quad CXR: Chest X-Ray; ET: enhancing tumor; TC: tumor core; WT: whole tumor; AVG: average; CF: Cerebrospinal Fluid; GM: Gray matter; WM: White matter; VS: Vestibular schwannoma.}
\label{table:hd95_2}
\end{table}